\documentclass{article}
\usepackage{ijcai23}

\usepackage{times}
\usepackage{soul}
\usepackage{url}
\usepackage{epstopdf}
\usepackage[hidelinks]{hyperref}
\usepackage[utf8]{inputenc}
\usepackage[small]{caption}
\usepackage{graphicx}
\usepackage{amsmath}
\usepackage{amsthm}
\usepackage{booktabs}
\usepackage{algorithm}
\usepackage{algorithmic}
\usepackage[switch]{lineno}
\usepackage{comment}
\usepackage{xcolor}
\definecolor{red}{rgb}{1, 0, 0}
\definecolor{blue}{rgb}{0, 0, 1}

\usepackage{array}
\newcolumntype{L}[1]{>{\raggedright\arraybackslash}p{#1}}
\newcolumntype{C}[1]{>{\centering\arraybackslash}p{#1}}
\newcolumntype{R}[1]{>{\raggedleft\arraybackslash}p{#1}}

\usepackage{caption}
\usepackage{subcaption}

\usepackage{multirow}
\usepackage{makecell}

\title{The Safety Challenges of Deep Learning \\ in Real-World Type 1 Diabetes Management}

\author{
Harry Emerson$^1$
\and
Ryan McConville$^1$ \And
Matthew Guy$^{1,2}$
\affiliations
$^1$University of Bristol\\
$^2$University Hospital Southampton\\
\emails
\{harry.emerson, ryan.mcconville\}@bristol.ac.uk,
matthew.guy@uhs.nhs.uk
}

\begin{document}

\maketitle

\begin{abstract}
Blood glucose simulation allows the effectiveness of type 1 diabetes (T1D) management strategies to be evaluated without patient harm. Deep learning algorithms provide a promising avenue for extending simulator capabilities; however, these algorithms are limited in that they do not necessarily learn physiologically correct glucose dynamics and can learn incorrect and potentially dangerous relationships from confounders in training data. This is likely to be more important in real-world scenarios, as data is not collected under strict research protocol. This work explores the implications of using deep learning algorithms trained on real-world data to model glucose dynamics. Free-living data was processed from the OpenAPS Data Commons and supplemented with patient-reported tags of challenging diabetes events, constituting one of the most detailed real-world T1D datasets. This dataset was used to train and evaluate state-of-the-art glucose simulators, comparing their prediction error across safety critical scenarios and assessing the physiological appropriateness of the learned dynamics using Shapley Additive Explanations (SHAP). While deep learning prediction accuracy surpassed the widely-used mathematical simulator approach, the model deteriorated in safety critical scenarios and struggled to leverage self-reported meal and exercise information. SHAP value analysis also indicated the model had fundamentally confused the roles of insulin and carbohydrates, which is one of the most basic T1D management principles. This work highlights the importance of considering physiological appropriateness when using deep learning to model real-world systems in T1D and healthcare more broadly, and provides recommendations for building models that are robust to real-world data constraints.
\end{abstract}

\section{Introduction}

% Introduce HCLS and T1D briefly
Hybrid closed loop systems (HCLS) reduce the management burden for people with type 1 diabetes (T1D); providing a means of automatically monitoring and regulating blood glucose levels with less patient interaction and cognitive burden. These devices have had a transformative effect on T1D management and are associated with a reduced risk of dangerous low blood glucose events and a longer time spent in the target glucose range \cite{Leelarathna2021Hybrid2021c}. HCLSs are composed of an insulin pump and a continuous glucose monitor (CGM) linked by a control algorithm. The CGM estimates the user's blood glucose concentration through regular measurements of blood glucose in the interstitial fluid and the control algorithm instructs the insulin pump to infuse background (basal) insulin to lower blood glucose to the desired level. HCLSs are not completely autonomous, for example, requiring additional carbohydrate information to calculate mealtime (bolus) insulin to compensate for the rise in blood glucose that occurs post carbohydrate consumption. 

% Discuss in-silico validation for clinical trials
Blood glucose control algorithms are an active area of diabetes research \cite{Kesavadev2020TheReview,Moon2021CurrentEvidence,Tejedor2020ReinforcementReviewb}, with efforts focused on developing safer and more intelligent control methods. To be suitable for patient use, newly developed algorithms must undergo rigorous evaluation in clinical trials and demonstrate benefit to the patient even in extreme conditions \cite{Smaoui2020DevelopmentAlgorithms}. Computer simulation provides an opportunity in T1D to accelerate regulatory approval and identify potential safety issues before moving towards human testing \cite{Blauw2016}. In addition, in-silico trials are also faster to perform, more cost-effective and enable the evaluation of patient groups and scenarios that would otherwise be unethical in a real-world setting \cite{Fritzen2018ModelingImpact}. These simulators are typically based on non-linear systems of ordinary differential equations and describe plasma-glucose dynamics through modelling the absorption of glucose and insulin between interconnected metabolic compartments in the body \cite{Nath2018PhysiologicalReview}. 

% Limitations of current simulator models
Despite the clinical successes of T1D simulators, these systems still represent a simplification of true blood glucose dynamics, with mathematical simulation models unable to accurately fully incorporate important factors such as exercise and stress \cite{Smaoui2020DevelopmentAlgorithms}. Furthermore, mathematical models are usually derived from patient studies of metabolic processes and validated on a limited number of highly-standardised clinical trial datasets. This means that updating simulator dynamics to incorporate newly observed phenomena often requires lengthy periods of patient data collection and that the resulting simulations are not necessarily representative of the wider T1D community \cite{Bhonsle2020AStrategies}. 

% Introduce machine learning
The growing application of machine learning in diabetes \cite{Woldaregay2019Data-drivenDiabetesc}, represents an opportunity for deep learning models to extend the current state-of-the-art in simulation and learn detailed models of blood glucose dynamics from real-world, more-readily available samples of patient data. In this context, a neural network is used to build a generalisable model of glucose dynamics and predict a patient's future blood glucose in response to different insulin dosing strategies. Deep learning, in particular, constitutes a state-of-the-art approach in blood glucose forecasting and is currently undergoing trials in patient populations \cite{Porumb2020PrecisionECG,Cescon2021ActivityConditions}. 

Despite their well-established prediction accuracy, there has been limited research into the physiological appropriateness of the learned dynamics of deep learning models and the implications this may have on the safety of real-world patients. This disparity has been observed in other areas of healthcare and  arises from unrecorded information influencing the training data \cite{Zhao2020TrainingApplications}, resulting in the model exhibiting low residual test error, while not demonstrating to have learned the correct dynamics. 

% Introduce the remainder of the paper
This work presents a detailed analysis of the challenges of using deep learning methods to generate physiologically appropriate predictions of blood glucose dynamics. The state-of-the-art glucose forecasting algorithm, Dilated RNN \cite{Zhu2022PersonalizedMeta-Learning} is used to represent a broader class of deep learning simulators. Prediction error with respect to the blood glucose values recorded in the dataset was then compared to several baseline simulators incorporating varying degrees of experimentally-derived diabetes knowledge. This includes the neural network based method introduced by Kushner et al. \cite{Kushner2020Multi-HourModels-correct}, in which the effects of insulin and carbohydrates are constrained to align with basic metabolic principles, and the Hovorka mathematical model \cite{Hovorka2004NonlinearDiabetes}, which is one of most-widely used simulators in T1D research. This comprehensive evaluation is performed using data from the real-world and publicly-available OpenAPS Data Commons and identifies the challenges of using more widely available and representative non-clinical trial data for training deep learning simulators. This work seeks to demonstrate the potential for deep learning to create more realistic simulations, while highlighting the limitations of state-of-the-art blood glucose forecasting models in learning physiologically appropriate relationships. The main contributions of this work are as follows: 

\begin{itemize}

    \item \textbf{Evaluation of deep learning in common glucose control scenarios with real-world data.} Important control scenarios, such as exercise, high fat meals and alcohol consumption, are identified from the OpenAPS data commons and used to probe the prediction error of deep learning based blood glucose models. The deep learning method achieves significantly reduced prediction error compared to the widely-used Hovorka mathematical simulator, but shows high uncertainty in safety critical scenarios. 

    \item \textbf{Comparison of learned blood glucose dynamics to experimentally-validated relationships.} Shapley Additive Explanations were used to interpret the learned blood glucose dynamics of the deep learning model. Analysis identified that even fundamental insulin-carbohydrate relationships were sometimes confused by the deep learning model, highlighting the need for methods to incorporate expert knowledge and for evaluation metrics which consider the physiological appropriateness of predictions, in addition to test dataset prediction error. 

    \item \textbf{Time-series free-living dataset for blood glucose forecasting and type 1 diabetes analysis tasks.} This work extracts blood glucose, insulin and carbohydrate information from the publicly-available OpenAPS Data Commons and processes them to form a comprehensive time-series dataset. Patient self-management notes were also processed, assigning specific labels of the important blood glucose control events exercise, high fat meals, high protein meals, alcohol and caffeine. The resulting dataset provides a rich data source for developing future T1D decision-making and forecasting tools. The code for the work is available here: \href{https://github.com/hemerson1/OpenAPS_Cleaner}{https://github.com/hemerson1/OpenAPS\_Cleaner}.

    \item \textbf{Feasibility of using self-reported labels for modelling physical activity and high fat meal consumption.} This preliminary work demonstrates that qualitative self-management notes of meal content and activity are ineffectual in improving the deep learning modelling of exercise and fatty meals events. This insight may suggest greater detail is required in patient note-taking to accurately model these high-risk validation scenarios or that a larger number of labelled events are required to effectively impact modelling.  

    \item \textbf{Case study on the challenges of using real-world data for deep learning in healthcare tasks.} This work highlights features of free-living data which can lead to dangerously incorrect treatment in type 1 diabetes, such as unreported carbohydrate consumption or high correlation between insulin and carbohydrates. These insights extend beyond their applicability to diabetes, as many healthcare domains utilising real-world training data are likely to face similar challenges stemming from unreported or highly correlated variables. As a result, the findings of this study will inform dataset and model design in broader healthcare contexts.
    
\end{itemize}

\section{Related Work} 

% Introduce mathematical simulators
Mathematical models of glucose dynamics represent the most common and established method of evaluating novel insulin dosing strategies. Of those available the most widely used are the Sorenson, UVA/Padova and Hovorka model \cite{Pompa2021ASystem}. The Sorenson model represents potentially the most sophisticated of the three models; using 22 non-linear equations and 135 patient parameters to model the evolution of glucose concentration in major organs such as the brain, liver, tissue, muscles and gut \cite{Panunzi2020ALoad}. The UVA/Padova simulator is the only simulator of blood glucose dynamics approved by the FDA as a substitute for animal trials in control algorithm development. This model describes glucose-insulin dynamics in people with T1D and can replicate intra-day insulin sensitivity variation and model the effect of different administration routes on insulin activity \cite{Visentin2018TheDayb}. The Hovorka model represents the simplest simulator of the described approaches; providing a fundamental description of gastro-intestinal absorption and subcutaneous insulin administration \cite{Pompa2021ASystem}. In addition, more fundamental models of glucose-insulin dynamics have been extended to model the effect of physical activity on insulin-glucose dynamics \cite{Alkhateeb2021ModellingDiabetes,Resalat2019AModel,Romeres2021ExerciseStudy}.   

% Discuss deep learning approaches in T1D simulation
A search of prominent academic databases including PubMed, IEEE Xplore, and Google Scholar of the search terms ``type 1 diabetes", ``deep learning", ``simulation" and ``blood glucose" returned a single example in which deep learning simulation had been explicitly used for evaluating blood glucose control algorithms in T1D. Deng et al. applied a systems biology informed neural network approach to build an individual model of a patient's blood glucose dynamics \cite{Deng2022Patient-specificDiabetes}. This approach utilises a hybrid approach, in which a mathematical model informs the basic dynamics of the simulator and updates the parameters of both a neural network and the mathematical model to minimise the error between the true and predicted blood glucose values \cite{Yazdani2020SystemsDynamics}. 

% Deep Learning in T1D Prediction
Adjacent to simulation, blood glucose forecasting requires an understanding of glucose dynamics. Deep learning is well-utilised for blood glucose forecasting with many approaches seeking to improve model accuracy using increasingly sophisticated model architectures and additional sensor modalities \cite{Woldaregay2019Data-drivenDiabetesc,Zhu2021DeepReview,Ahmed2023TheReview}. Notable examples of deep learning in blood glucose forecasting include: GluNet, an extension of the successful dilated CNN time-series prediction method \cite{Li2020GluNet:Forecasting}, PolySeqMO, which applies a recurrent decoder method to infer the parameters of an \(n\) degree polynomial for multi-horizon prediction \cite{Fox2018DeepForecastingb}, and a GAN approach presented by Zhu et al. in which the model generator makes predictions of future blood glucose and the discriminator improves prediction similarity with patient data \cite{Zhu2020BloodNetworks}. Forecasting algorithms have also sought to combine mathematical and deep learning techniques in hybrid simulator approaches, allowing for both adaptive and physiologically meaningful blood glucose predictions. For example, Miller et al. applies a deep state space model to dynamically vary the parameters of the UVA/Padova model in response to contextual information \cite{Miller2020LearningWildb}. 

% Evaluation of deep learning in prediciton
Outside of the exploration covered in this work, Hameed et al. compared the performance of several deep learning models trained on the free-living OpenAPS Data Commons and the clinical OhioT1DM dataset \cite{Hameed2020ComparingData}. This evaluation differs from the approach presented in this work as analysis focused on identifying machine learning algorithms with improved accuracy across the two data settings, as opposed to determining if learned blood glucose dynamics were correct and representative of real patient dynamics. Similarly, Kushner et al. presented a method for verifying the correctness of the insulin dynamics of deep learning algorithms \cite{Kushner2020ConformanceDynamics}. Their deep learning approach was trained on clinical trial data and highlighted the high test set accuracy of the approach, despite the development of incorrect insulin dynamics. This paper builds on the work of Kushner et al., extending the analysis to a much larger and more detailed real-world dataset. Furthermore, this work utilises Shapley Additive Explanations to describe the contribution of individual sample features, modelling insulin and carbohydrate dynamics over a considerably longer time period to create a more detailed and holistic view of the learned model dynamics.   

% Offline Evaluation in Deep Learning
In the wider field, deep learning is an established approach for evaluating algorithmic performance in offline reinforcement learning, in which a machine learning driven agent learns the action sequence to maximise reward in a dynamic environment by observing demonstrations in a static dataset \cite{Prudencio2023AProblems}. This application of deep learning for algorithmic evaluation is commonly referred to as off-policy evaluation and approaches typically utilise one of: Q-value estimation, in which a neural network is used to approximate the expected total reward of an action in a given scenario, importance sampling, in which dataset rewards are weighted based on their likelihood of occurring under the evaluation algorithm, or by a model-based approach, in which environmental dynamics are approximated via a neural network \cite{Fu2021BenchmarksEvaluationb}. Off-policy evaluation is invaluable in settings where evaluating in the dynamic environment would be costly or potentially dangerous, therefore it has been applied to a variety of topics in healthcare such as anesthetic administration \cite{Cai2023TowardsLearning}, structuring chemotherapy regimens \cite{Shiranthika2022SupervisedLearning} and optimising ventilation treatment \cite{Kondrup2022TowardsLearning}. 

\section{Methods}

\subsection{Processing the OpenAPS Data Commons}

% Introduce the OpenAPS dataset
The OpenAPS Data Commons represents one of the most extensive collections of free-living T1D data and is composed of data from 184 people across 23 countries (\href{https://openaps.org/outcomes/data-commons/}{https://openaps.org/outcomes/data-commons/}). Participants in the repository predominantly utilise do-it-yourself (DIY) HCLS devices, which are community-designed insulin delivery systems built using commercial CGMs and insulin pumps. User data was donated to the project and not collected within a clinical trial setting, therefore the data is vast and inconsistent between patients. Each participant at a minimum uploaded blood glucose measurements recorded by their CGM, however there is otherwise considerable variation in insulin delivery system, data collection period, device models, metrics logged and patient management strategy. Data of this form is typically automatically logged and uploaded by HCLSs and consequently is in greater abundance than the formal study data used for training most machine learning models in T1D \cite{Woldaregay2019Data-drivenDiabetesc}. In addition, the data quality is much more representative of the quality in real-world patients, as participants in the repository are not adhering to strict data collection protocols.   

To convert the dataset into a suitable form for training machine learning models, a subset of the participants were selected based on the following criteria: 

\begin{itemize}

    \item \textbf{HCLS User} - Participants must have used an HCLS for insulin dosing at some point in the dataset. These systems continually optimise insulin infusion rates, creating a diverse dataset of insulin actions for training machine learning models. 

    \item \textbf{AndroidAPS Software} - Participants must be using an HCLS incorporating the AndroidAPS app. There are several common open-source software packages used for automated insulin dosing in HCLSs (e.g. AndroidAPS, OpenAPS, Loop) \cite{Jennings2020Do-It-YourselfProfessionals}. However, AndroidAPS is the only platform present in this dataset which logs all the insulin doses delivered by the system, including the default basal rates which are, in effect, outside of control algorithm activity. 
    
    \item \textbf{Available Demographic Data} - Participants must have demographic data recorded in the OpenAPS Data Commons such as their weight, age, mean total daily insulin and mean carbohydrate consumption. This data allows commonalities to be identified between similar patients in the dataset and is helpful for training a machine learning model capable of generalisation. 
    
\end{itemize}

% Explain differences with existing method
The resulting subset contained 18 applicable participants, with the majority removed for using non-AndroidAPS software.
This subset was selected  for machine learning models to learn correct insulin-glucose relationships. Default basal infusion rates are commonly set at half-hour graduations by the user and are enacted outside of control algorithm activity. These insulin doses make up approximately 50\% of all insulin actions in the dataset and consequently have a considerable influence on participant blood glucose dynamics. The single prior work utilising a larger sample of OpenAPS participants applied forward-filling between successive basal rates recorded in the dataset \cite{Hameed2020ComparingData}. This may act as a suitable approximation for the default basal rates when unavailable, but will ultimately impede learning blood glucose dynamics that are representative of real-world patients. 

% Summarise the resulting dataset
In the broader T1D community, AndroidAPS is the most common software package for DIY HCLS users, therefore it should be feasible to expand to larger patient cohorts in future work \cite{Street2021ReviewOutcomes}. The remaining dataset subset contains approximately 1.6 million samples, spanning a combined 19 years of data collection. Of the available participants, the mean age was 36.2 (5.5 - 57.2) years and the mean quantity of uploaded data was 386 (6 - 1215) days. In addition, 11\% of the participants identified as female and 100\% identified as white. The mean HbA1c measurement at the time of data upload was 48 (36 - 62) mmol/mol. HbA1c provides a metric for measuring blood glucose control, with lower values indicating a lower mean blood glucose level. A multi-national study of mean HbA1c measurements concluded that most people with T1D do not achieve an HbA1c less than 58 mmol/mol \cite{Mcknight2015GlycaemicComparison}, therefore the improved control in this population may relate to the fact that HCLS technology is highly beneficial for blood glucose management.  

% Making continuous blood glucose segments
The filtered dataset samples were converted to discrete five minute time points by rounding the timestamp of each measurement to the nearest five minute interval. This frequency matches the CGM sampling rate of the devices in the OpenAPS Data Commons and was performed as the prediction methods utilised in this work were incompatible with irregularly-spaced temporal data. Although insulin dose timing has a considerable impact on blood glucose dynamics, altering the timing by a maximum of 2.5 minutes is unlikely to have an effect on the learned dynamics. In each five minute period, the most recent basal infusion rate was selected as the rate for that period; reflecting the fact that more recent insulin delivery instructions override prior actions. Temporary basal rates were accompanied by a duration and therefore the basal rate for every timestep in that period was set to the corresponding basal value. Bolus insulin and carbohydrates were summed for each interval as they represent isolated events rather than periods of ongoing activity. Missing measurements are not uncommon in CGM devices \cite{Lin2019HandlingRecords}, therefore logarithmic interpolation was applied between consecutive blood glucose values to fill gaps between recordings. Timesteps without bolus insulin doses or carbohydrates consumption values were set to zero to indicate that no action had occurred in that timestep. Similarly, entries without a corresponding basal rate were set to the default basal value for that time of day.

% Discuss addition of qualitative device data
In addition to the passively recorded device data, the OpenAPS Data Commons also included qualitative descriptions of blood glucose control events for the benefit of retrospective user analysis. Due to the non-standardised method by which individual users self-reported events, identification was performed manually and focused on labels of exercise, high fat meals, high protein meals, caffeine and alcohol. These events were selected as they have a well-established effect on blood glucose dynamics and could be feasibly identified from user notes \cite{Charlton2020ABeverages,Paterson2015TheManagementb,Paterson2016InfluenceTherapy,Abdou2021EffectEgyptb,Dewar2017TheDiabetes}. 

% Criteria for identifying events
Exercise events were identified via reference to physical activity, such as walking, cycling or running. Events describing food composition were identified by cross-referencing meal descriptions with the nutritional values of generic food items in the FNDDS dataset \cite{Montville2013USDA5.0}. If the food item contained more than 12.5 g protein, 20 g fat, 1 g alcohol or 1 mg caffeine per 100 g then the event was labelled accordingly. The thresholds of protein and fat were selected as these are the smallest quantities that have been observed to affect blood glucose levels \cite{Smart2020InsulinTime}. A degree of interpretation was required in selecting the correct reference for a given meal label, therefore in instances where multiple nutritional references were available meal items were classified based on the reference with greatest percentage composition. Furthermore, in cases of ambiguity, such as when a note references a restaurant chain, best judgement was used to determine the suitability of the food item for the aforementioned categories. Each labelled event was represented by a separate channel in the pre-processed dataset with a binary indicator of whether the event occurred in the prior five minute measurement period. From the described process approximately 6600 labels were identified covering 11 of the 18 participants (70\% exercise, 20\% high fat meals, 6\% high protein meals, 2\% caffeine consumption and 2\% alcohol consumption). 

\subsection{Blood Glucose Dynamics Simulation}

% Discuss the baseline methods
Simulation methods were selected to represent the three broad categories of glucose prediction methods: deep learning, mathematical, and hybrid. The dilated RNN algorithm presented in Zhu et al. was used to represent solely deep-learning based approaches \cite{Zhu2022PersonalizedMeta-Learning}. This algorithm is a competitive blood glucose prediction method and extends the established recurrent neural network with skip connections to aid in the learning of multi-resolution dependencies \cite{Chang2017DilatedNetworks}. This method was compared to the hybrid algorithm presented in Kushner et al. \cite{Kushner2020Multi-HourModels-correct} and the Hovorka mathematical dynamics model \cite{Hovorka2004NonlinearDiabetes}. The hybrid method utilised a single-layer, physiologically informed neural network such that the effect of insulin and carbohydrate on future blood glucose aligned with known relationships in T1D (i.e carbohydrates contribute positively and insulin contributes negatively) \cite{Kushner2020Multi-HourModels-correct}. In addition, the Hovorka model was also utilised as it represents one of the most widely-used methods of blood glucose simulation and requires relatively few parameters for model personalisation \cite{Hovorka2004NonlinearDiabetes}. 

%%%%%%%%%%%%%%%%%%%%%%%%%%%%%%%%%%%%%%%%%%%%%%%%%%%%%%%%%%%%%%%%

\begin{table}[h!]
\footnotesize
\centering

% Vertical and Horizontal Padding
\setlength{\tabcolsep}{0.5em}
\def\arraystretch{1.5}

\small
\begin{tabular}{L{2.5cm} L{5cm}}
\toprule
Metric & Description \\
\midrule

1) Blood Glucose & The most recent CGM measurement. \\

2) HCLS Insulin Infusion & The most recent basal rate set by HCLS control algorithm or participant.  \\

3) Total Insulin Infusion & The sum of all insulin taken since the previous timestep. \\

4) Insulin Activity & The IOB metric described in Zhu et al. \cite{Zhu2022PersonalizedMeta-Learning}, the approximate amount of active insulin in a participant's body. \\

5) Carbohydrate Consumption & The sum of all carbohydrates consumed since the previous timestep. \\

6) Carbohydrate Activity & The COB metric described in Zhu et al. \cite{Zhu2022PersonalizedMeta-Learning}, the approximate amount of active carbohydrates in a participant's body. \\

7) Mean Basal Rate & The mean default basal infusion rate as set by each participant. \\

8) Patient Weight & The participants self-reported weight at the time of data upload. \\

9) Time of Day & The hour of the day in which the timestep occurred. \\

\bottomrule
\end{tabular}
\caption{The features used for training the deep learning and hybrid simulator models. Model inputs also incorporated a binary indicator of exercise, high fat meals, high protein meals, alcohol consumption and high caffeine consumption where specified in text. 7), 8) and 9) are used to encourage knowledge transfer between participants. The mean basal rate gives an overview of participant's aggregate basal dosing behaviour, body weight is indirectly used to calculate bolus \protect\cite{Kuroda2012Carbohydrate-to-insulinPump} and time of day is used to identify intra-day patterns in blood glucose dynamics.}
\label{tab:input_features}
\end{table}

%%%%%%%%%%%%%%%%%%%%%%%%%%%%%%%%%%%%%%%%%%%%%%%%%%%%%%%%%%%%%%

% Discuss changes in implementations from original
Model architectures were kept consistent with their original implementations, except where necessary to introduce standardisation between methods. To this extent the loss function for the deep learning and hybrid approach was substituted for the glucose-specific mean-squared error (gMSE) presented in Favero et al. \cite{Favero2012AModels}. This loss function modified the standard mean squared error to include an additional penalty for incorrect predictions at potentially dangerous blood glucose values. This consisted of a penalty for overestimating blood glucose near the low blood glucose region and underestimation near the high blood glucose region. The input features of the models were also standardised across the deep learning and hybrid approaches using a modified version of the representation presented in Zhu et al. \cite{Zhu2022PersonalizedMeta-Learning}. The input features were modified to ensure sufficient demographic information was available for knowledge transfer between similar patients. Table \ref{tab:input_features} provides a detailed summary of the utilised features. 

% Breaking into 4-hour trajectories
Hyperparameters were also altered from their default values and tuned to each training dataset using Bayesian optimisation. To model the temporal dependencies of blood glucose dynamics, the input data was divided into overlapping four-hour segments. This interval was selected as the impact of insulin and carbohydrates on blood glucose is greatest during this period \cite{Cengiz2016MovingPumpsb}. Segments were excluded if a gap of more than 30 minutes was present between consecutive, non-interpolated blood glucose measurements. 

% Changing the Time horizon
The prediction horizons of the deep learning and hybrid models were also modified from their original implementations. In their use as blood glucose forecasting algorithms, prediction horizons of at least 30 minutes were utilised to allow patients sufficient time to intervene in predicted low blood glucose events. As the aim of this work was to investigate the correctness of the learned dynamics of deep forecasting models, the prediction horizon was also set to match the CGM sampling rate of five minutes. This decision was made to ensure that simulators had a full awareness of all the diabetes-relevant actions occurring between the current time and the time of prediction. For example, during the 30 minute prediction horizon the patient could take a large dose of insulin. However, the simulator model would have no awareness of this action and therefore would incorrectly attribute the change in blood glucose that might result from this to actions occurring prior to the current time. 

\subsection{Deep Learning Simulator Evaluation}

Experiments were designed to create a rigorous evaluation of the effectiveness of deep learning models in simulating blood glucose dynamics. This was performed with a focus on model predictions achieving low error with respect to the test dataset, aligning with known blood glucose relationships, and being applicable to patients across a wide range of scenarios. 

\subsubsection{i) Residual Error of Deep Learning Predictions}

Machine learning models were trained using the first 90\% of each participant's data, with the remaining samples being divided evenly for the purposes of validating model hyperparameters and testing the fully-trained configurations. Test and validation data were composed of the most recent samples collected under each participant. This was done to prevent information leakage between the various sets and present a more realistic evaluation. Evaluation was performed by comparing simulator predictions to the next blood glucose value in the processed dataset. Performance was assessed using the aforementioned gMSE function \cite{Favero2012AModels} and the significance of the ranking was confirmed via the Friedman rank test between the deep learning and benchmark approaches. 

To better identify model limitations, test scenarios were selected representing challenging blood glucose modelling tasks: meal consumption (non-zero carbohydrate consumption), night time (recorded between 9pm and 4am), high blood glucose (blood glucose measurement above the typical recommended range of 70 to 180 mg/dl) and low blood glucose (blood glucose measurement below 70 mg/dl). The night time range of 9 pm to 4 am was chosen as the majority of participants did not consume meals during this period and were, therefore, more likely to be asleep. These scenarios were selected in addition to those highlighted by the qualitative event labels of exercise, high fat, high protein, alcohol and caffeine. Despite the use of qualitative labels for identifying test scenarios, it is worth noting these events were not included as features for the model in this section of the work and are explored separately in experiment iii).

\subsubsection{ii) Physiological Appropriateness of Learned Blood Glucose Dynamics}

Low prediction accuracy is only useful if the blood glucose dynamics learned by the model are representative of the relationships observed in real-world patients. To verify model correctness, the predictions of the deep learning model were compared with experimentally observed blood glucose dynamics relationships. This was performed primarily by calculating the feature importance of the simulator inputs presented in Table \ref{tab:input_features}. Feature importance was determined using Shapley Additive Explanations (SHAP) and calculated via an extension of the method presented in Sundarajan et al. \cite{Sundararajan2017AxiomaticNetworks}. This method was selected as it provided a computationally efficient approach for approximating the SHAP values of a large neural network. The SHAP model was calibrated and tested on 1,000 blood glucose trajectories randomly sampled from the processed dataset. This sample size was selected to reduce the variance of the SHAP value estimates, while minimising the computational cost of calculation. The learned blood glucose impact of insulin dosing and carbohydrate consumption was then compared to the theoretical insulin-carbohydrate activity modelled within the Hovorka simulator. The SHAP method assumes independence between model input features, therefore to satisfy this constraint the insulin and carbohydrate activity features were removed for this experiment. 

This method was also extended to identify if simple dataset augmentations were sufficient for improving the learned blood glucose dynamics of the deep learning model. This exploration was motivated by a need for techniques to reinforce experimentally-validated glucose relationships without constraining model outputs and limiting model complexity. This was performed with a particular focus on unreported carbohydrate consumption, as this has been previously identified as a substantial challenge for developing glucose forecasting algorithms \cite{Kushner2020Multi-HourModels-correct}. The selected augmentations included:
\begin{itemize}
    
    \item \textbf{Filtering Outliers} - Remove training samples occurring on days where carbohydrate intake deviates from the interquartile range. This is under the assumption that on days with low carbohydrate intake unreported meal consumption has occurred. After dataset filtering, the total number of glucose measurements in the dataset was reduced from 1.6 million to 1.1 million.

    \item \textbf{Labelling Unreported Carbohydrates} - Label unannounced carbohydrate consumption using the method presented in Zheng et al. \cite{Zheng2019AutomatedExplanationb}. This method involved using the mathematical Hovorka simulator to identify blood glucose trajectories in the dataset that diverge considerably from simulator predictions. At divergence points, a grid-based method was used to select the optimal carbohydrate quantity and timing for the unreported meal. This increased the mean frequency of carbohydrate consumption from 5.34 to 6.65 meals per day (including snacks above 15 g of carbohydrates).     

\end{itemize}
The learned dynamics achieved under the modified datasets were then compared to those obtained under the unaltered implementation. Dynamic time warping was used to assess the overall error between learned and theoretical dynamics. This metric was chosen as it accounts for delays in time-series data and has been previously applied to measure similarity between glucose trajectories \cite{Yu2022DeepDiabetes}. The significance of the dynamics changes were verified via Friedman rank tests, in which dynamics under the two augmentations were compared to the unaugmented dataset.  

\subsubsection{iii) Using Self-Reported Labels to Model Complex Glucose Behaviour}

% Overview of label selection
Using self-reported labels of physical activity and meal composition may provide a simple method for modelling complex glucose dynamics behaviours unavailable in the current mathematical simulators. To explore this, the methodology described in i) was extended to include the following modifications. The input features described in Table \ref{tab:input_features} were supplemented with binary labels of exercise and high fat meal consumption. These events were selected as they represented 90\% of all labelled events and each provided at least 500 demonstrations in the processed dataset. Instead of using the full cohort as in prior experiments, a single participant was selected for training and testing purposes. This participant possessed approximately 80\% of the total event labels and hence had consistent sample coverage over the full data collection period.

% Summary of labelled dataset statistics
To ensure there were sufficient testing scenarios available, the test size was increased to 45\% of participant's data with the training set reducing to 50\%. This resulted in the final training set containing approximately 150,000 samples with 2,500 scenario labels; equating to approximately 1.5 years of data with a mean 4 exercise events per day and 1 high fat meal every other day. The modified simulator was trained as described in experiment i), with each model's hyperparameters being tuned to the single participant's dataset and significance . Evaluation was performed as before, with comparisons made between the deep learning model with and without self-reported event labels and statistical significance being measured using a Wilcoxon-signed rank test.    
  
\section{Results}

\subsubsection{i) Deep Learning Simulators Achieve Lower Prediction Error, but Struggle in Safety Critical Scenarios}

\begin{figure*}[h]
    \centering

    \hspace*{-0.8cm}
    \includegraphics[width=0.95\textwidth]{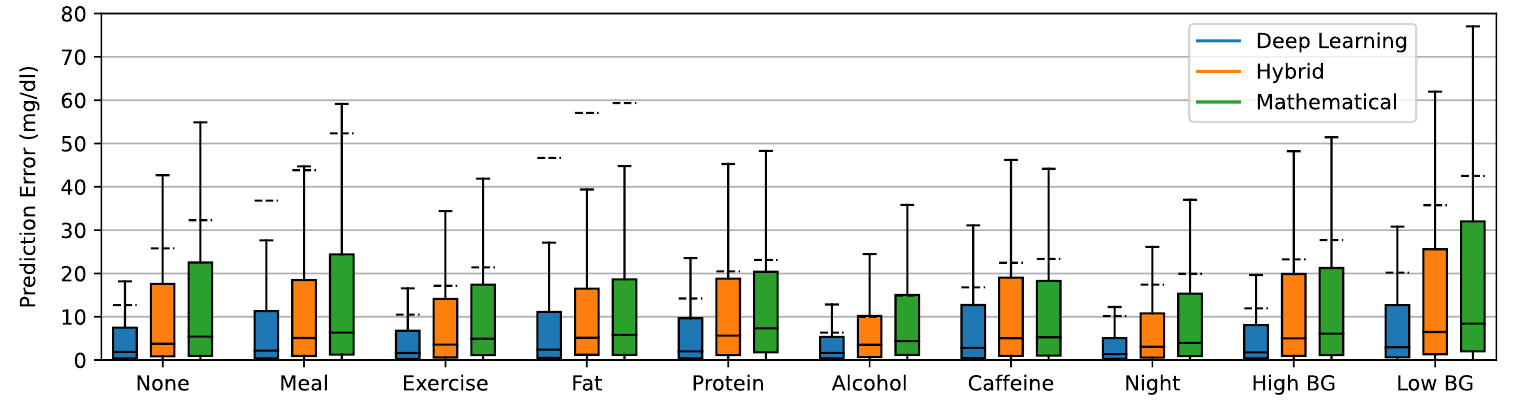}
    \caption{\textbf{Comparison of the different blood glucose dynamics simulation methods in common control scenarios.} The deep learning simulator can be seen to yield lower residual error compared to the mathematical and hybrid model approach. Greater uncertainty is obtained in high-risk scenarios such as meals, high fat meals and low BG (blood glucose) events as evidenced by the larger range. Statistical significance was assessed using the Friedman rank test across all test scenarios (p \(<0.001\)). The scenarios labelled None, Meal, Night, High BG and Low BG were all validated on 500 samples. Exercise, Protein and Fat on 100 samples and Alcohol and Caffeine on 20 samples. The central black line represents the median ad the dotted black line, the mean. The whiskers extend to 1.5 times the interquartile range.}
    \label{fig:full_comparison}
\end{figure*}

% How does the accuracy compare between models?
Figure \ref{fig:full_comparison} details the performance of the described simulator methods across a series of important test scenarios in the validation of T1D control algorithms. From the figure it is clear the performance of the deep learning algorithm exceeds baseline simulator methods, 
achieving a lower median prediction error and reduced variability across all the control scenarios (a median error reduction of \(1.9 \pm 0.9\) mg/dl compared with the hybrid approach and \(3.5 \pm 0.9\) mg/dl with the mathematical model). The increased accuracy of the deep learning approach highlights the potential of the method for learning complex glucose behaviour from real-world data, demonstrating reasonable improvements in common high-risk control scenarios, such as meals and exercise.

% Justifying that high accuracy is not always positive
The low residual error of the deep learning approach does not necessarily suggest learned blood glucose dynamics are representative of real-world patient dynamics and could feasibly result from the simulator adapting to samples influenced by missing or misreported data. For example, when a participant corrects for a low blood glucose event they are typically advised to ingest 15 to 20 grams of short-acting carbohydrates without any corresponding insulin dose \cite{Briscoe2006HypoglycemiaManagement}. Whilst carbohydrate information is typically provided when calculating an insulin bolus, many users do not let the HCLSs know of carbohydrates given for low blood glucose treatment (even though the facility is there for them to do so), meaning low blood glucose correction meals are often left unrecorded. A simulator bound by metabolic principles could be expected to perform poorly in these scenarios as it is lacking important carbohydrate information relevant to modelling the scenario correctly. In contrast, the deep learning method is not subject to any such constraints and would adapt to the physiologically inappropriate relationships and achieve lower prediction error relative to the test dataset. The observed performance of the three simulator approaches suggests this may be the case for the processed dataset, as the physiologically constrained hybrid and Hovorka models achieve their greatest median error of \(6.5 \pm 0.7\) mg/dl and \(8.4 \pm 0.8\) mg/dl in low blood glucose events specifically and also exhibit particularly large prediction uncertainty. In contrast, the deep learning approach exhibits its largest reduction in median error of \(3.5 \pm 1.0\) mg/dl and \(5.5 \pm 1.2\) mg/dl, respectively, suggesting the approach may be adapting to unrealistic aspects of the scenario observed within the dataset, such as the presence of unreported low blood glucose correction meals.

% Which scenarios does the data-driven model perform poorly on?
High uncertainty across the three simulation approaches was also obtained in the meal and high fat meal scenarios with an interquartile range of \(10.9 \pm 6.2\) mg/dl and \(10.7 \pm 6.0\) mg/dl, respectively. Furthermore, the mean prediction error for these events was far in excess of the median, suggesting the presence of very high error outliers. This disparity could potentially arise from unreliable carbohydrate information that may be present in the OpenAPS Data Commons. People with T1D must estimate the carbohydrate content of the meals they consume and take a bolus insulin dose proportional to the quantity. Carbohydrate calculation is a challenging task and estimation errors are common, with patients in prion work being observed to overestimate carbohydrates by a mean of 40\% \cite{Meade2016AccuracyAdults}. This inherent uncertainty may justify the increase in prediction error observed in this scenario. Probing the relationship between prediction error and a sample's carbohydrate activity demonstrates a weak positive correlation between larger carbohydrate meals and greater prediction error. With a Spearman rank correlation of 0.23 (p \(<\) 0.001) across 1,000 random dataset samples. This result may also support the assumption that greater meal inaccuracy results from carbohydrate calculation error, as a 40\% estimation error will result in a greater carbohydrate disparity for larger meals.  

% Uncertainty around fatty meal consumption
Similarly, dietary fat consumption alongside carbohydrates has been demonstrated to delay glucose absorption, resulting in a greater likelihood of prolonged blood glucose rises post-meal \cite{Smith2021InsulinReview}. As meal fat composition is not explicitly indicated by features of the training data, the simulator will likely experience greater uncertainty in these scenarios resulting from the hidden information. Despite the overall improvement observed with the deep learning simulator, the relatively high variance experienced in meal, high fat meal and low blood glucose scenarios is concerning for the approaches use in control algorithm validation and blood glucose forecasting. These events represent some of the most challenging T1D management scenarios and therefore low prediction error and high consistency would be essential for building trust in methods validated via machine learning based simulation. 

%%%%%%%%%%%%%%%%%%%%%%%%%%%%%%%%%%%%%%%%%%%%%%%%%%%%

\begin{figure*}[t!]
\begin{subfigure}[b]{0.5\textwidth}
    \centering
    \includegraphics[width=0.95\textwidth]{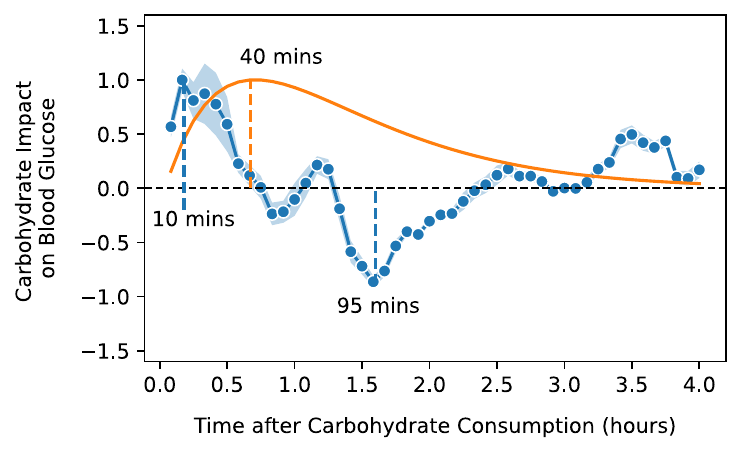}
    \caption{}
    \label{fig:carb_activity}
\end{subfigure}
\hfill
\begin{subfigure}[b]{0.5\textwidth}
    \centering
    \includegraphics[width=0.95\textwidth]{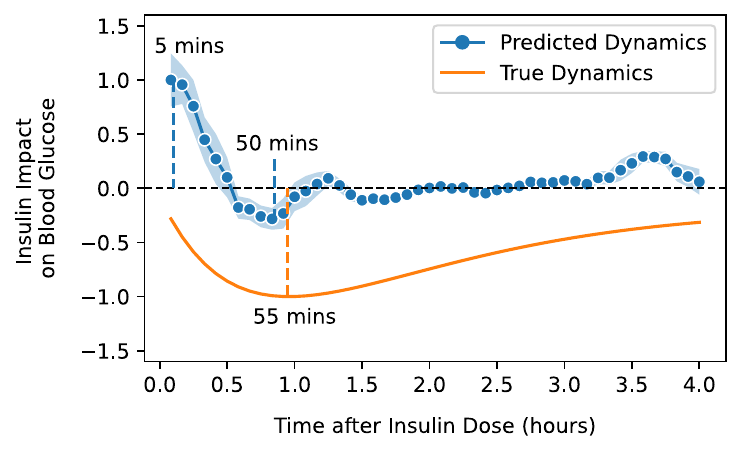}
    \caption{}
    \label{fig:ins_activity}
\end{subfigure}

 \caption{\textbf{Comparison of the learned and theoretical insulin-carbohydrate dynamics of the deep learning model.} a) The learned carbohydrate dynamics of the deep learning simulator compared to the theoretical dynamics of the Hovorka model. The deep learning model incorrectly concludes carbohydrates lower future blood glucose levels as suggested by the mean negative impact of carbohydrates 95 minutes after consumption. b) The learned insulin dynamics of the deep learning simulator compared to the theoretical dynamics of the Hovorka model. Insulin correctly shows a negative effect on blood glucose at approximately one hour after insulin dosing, but deteriorates considerably in the first 30 minute period and following the two hour point. Error bars indicate the standard error.}
\end{figure*}

%%%%%%%%%%%%%%%%%%%%%%%%%%%%%%%%%%%%%%%%%%%%%%%%%%%%

% Why is mathematical model performance poor?
Of the presented simulation methods, the mathematical model obtains the greatest median error across all test scenarios. This is unexpected as mathematical glucose dynamics simulators are widely used for blood glucose prediction in model predictive controllers \cite{Bequette2013AlgorithmsControl}. The performance deterioration may result from unreported factors influencing blood glucose dynamics, but could also arise from difficulty in calibrating the simulator for individual patients. The parameters of the Hovorka model can be tuned to patient blood glucose trajectories to obtain greater prediction accuracy, but the implementation can be simplified by calculating these parameters based on the patient's weight and setting the remaining values to their defaults \cite{Bequette2013AlgorithmsControl}. In this work the simplistic implementation was selected, as the machine learning simulator was not fine-tuned on individual patient data. The method presented in Miller et al. demonstrates the benefit of tuning mathematical model parameters on specific patients, whereby superior performance is achieved using the UVA/Padova simulator when dynamically updating the insulin sensitivity parameter \cite{Miller2020LearningWildb}. Further disparity may have resulted from uncertainty in participant weight measurements. This value was self-reported at the time of data upload and was not updated throughout the full data collection period. It is improbable that a participant would not exhibit a large percentage change in weight across multiple years of CGM data, therefore the parameter would likely be sub-optimal for at least a portion of the test set; resulting in lower accuracy predictions.  

\subsubsection{ii) Simple Insulin-Carbohydrate Relationships can become Easily Conflated in Real-World Data}

% Basic analysis of mean insulin impact and carbohydrate impact being wrong -> insulin-carbohydrate conflation
Figure \ref{fig:carb_activity} and Figure \ref{fig:ins_activity} demonstrate the learned blood glucose dynamics of the deep learning model. From the figures, it is evident the simulator has drawn incorrect conclusions about the effect of insulin and carbohydrates on blood glucose dynamics; with the results in direct disagreement with physiological observations. Figure \ref{fig:ins_activity} shows the model to have incorrectly learned that insulin has the overall effect of raising future blood glucose levels; with a net positive contribution of \(0.08 \pm 0.04\) across the full four-hour period. Similarly, Figure \ref{fig:carb_activity} shows two distinct negative troughs in carbohydrate impact at 50 minutes and 95 minutes, a substantial deviation from theoretical predictions. The low error of the deep learning model relative to the benchmark approaches in section i) suggests this relationship is unlikely to result from model underfitting and instead could be caused by the inability of the deep learning model to account for confounding variables in the dataset. In most HCLSs, carbohydrates are only recorded for the purpose of bolus insulin calculation. Similarly, individual insulin doses are often only given to account for carbohydrates. These two factors cause large insulin doses and carbohydrate consumption to be correlated within the dataset with a Spearman Rank Correlation of 0.32 (p \(< 0.001\)); making it potentially difficult for simulators to distinguish their individual contributions. Insulin-carbohydrate conflation is highly important for training future deep learning models, as the concept that carbohydrates raise blood glucose levels and insulin causes them to fall are two of the most fundamental principles of T1D management. A blood glucose simulator that is unable to identify this would be highly dangerous for control algorithm evaluation and development.

% Peaks at roughly the right places (high in carb activity at roughly at 10-25 mins vs 40 mins in theory, trough in insulin activity at 50 mins vs 55 in theory)
Neglecting insulin-carbohydrate confusion, greater insight can be gained by probing the peak impact times of the model. From examining the points of greatest impact in Figure \ref{fig:carb_activity}, it is evident that carbohydrate ingestion has the greatest positive effect on blood glucose approximately 10 to 25 minutes post meal. Similarly, in Figure \ref{fig:ins_activity}, insulin decreases blood glucose most notably 45 to 55 minutes after insulin dosing. Furthermore, a characteristic decline in insulin impact is observed following the peak. These results do approximately align with the theoretical peak carbohydrate and insulin activity times of 40 and 55 minutes, as modelled by the Hovorka model for a carbohydrate-only meal and a rapid acting insulin solution. This suggests that the model can at least partially describe the temporal features of insulin-carbohydrate dynamics. The comparably larger disparity between theoretical and learned peak carbohydrate impact could be attributed to inaccuracy in patient reporting of carbohydrate consumption. T1D management guidelines recommend insulin to be given in advance of carbohydrates to account for the lag in subcutaneous insulin action \cite{Slattery2018OptimalReview}. However, in practice, these events are typically recorded as having occurred simultaneously at the point of insulin infusion. In addition, small disparities between the learned and theoretical peak insulin impact time may result from participants in the dataset using different insulin solutions than those described by the Hovorka simulation or alternatively may be due to the intrinsic measurement delay of most commercial CGM devices.

% Spike in carbs in 5 mins -> Could be in response to blood glucose rising
Highly erroneous impact peaks are also present in the learned blood glucose dynamics, with insulin considerably raising blood glucose 5 minutes after infusion. The peak in insulin impact most likely results from the tight coupling of bolus insulin and meal carbohydrates, causing their effects to become conflated. Peak carbohydrate activity typically precedes peak insulin activity, therefore insulin and carbohydrates given at the same time oftens result in a characteristic rise and fall in blood glucose levels \cite{Akturk2018PossibleDiabetes}. To an observer without a knowledge of T1D management, it may appear that the insulin dose resulted in blood glucose rising, despite the timeframe being too short for it to have any reasonable effect.

% Significant trough in carbs at 95 mins -> Could this be associated with post-prand drop
Similarly, carbohydrate has the erroneous effect of considerably lowering blood glucose at 95 minutes. This drop in carbohydrate impact could feasibly align with the drop in blood glucose associated with the characteristic post-meal peak. The fall in blood glucose corresponds to the depletion of carbohydrate impact and is caused by there being more active insulin in the body at this point in time than is needed to account for the consumed meal. Both people with and without T1D experience a peak in blood glucose post-meal, which has been observed in people with T1D to occur approximately 57 to 100 minutes after meal consumption \cite{DenckerJohansen2012InterindividualMellitus}. As the trough in carbohydrate impact begins to form 70 minutes after meal consumption, this feature may result from the decline of the post-meal peak.  

% Peak in insulin and carbohydrate activity at approximately 210 minutes, why?
% Maybe insulin has run out at this point? 
A small positive impact peak is also present approximately 210 minutes after insulin infusion and carbohydrate consumption. This peak could potentially result from the diminishing of insulin activity before the entirety of the consumed carbohydrates have become metabolised. Most HCLSs are designed to use rapid acting solutions, such as Aspart and Lispro \cite{Berget2019ATherapy}, in which the duration of significant activity ranges between three to five hours based on a patient's physiology \cite{Wong2021Ultra-rapid-actingNeeded}. If a large number of dataset participants frequently eat high fat meals, which delays the effect of glucose absorption \cite{Smith2021InsulinReview} or are utilising ultra-rapid acting insulin, it may be challenging for a deep learning model to attribute a rise in blood glucose almost four hours post-meal to the depletion of a patient's meal insulin. This is accentuated by the model lacking knowledge of a meal's fat content and the activity duration of each patient's chosen insulin solution.   

% Word of warning about methodology of SHAP value explainability? 
It is worth noting that the method for calculating model SHAP values may be partly affected by the correlation between insulin and carbohydrates, as the approach does rely on the assumption of feature independence. This method was selected as it has been widely used for deep learning explainability across healthcare domains \cite{Smartphone-basedScienceDirect,Nohara2022ExplanationHospital} and has been applied to blood glucose forecasting evaluation in prior works \cite{Cappon2020ADiabetes}. Similarly, feature independence is rarely possible in a real-world data setting and therefore this method allows the approximate dynamics of the model to be probed \cite{Mase2019ExplainingRefinement}. 

%%%%%%%%%%%%%%%%%%%%%%%%%%%%%%%%%%%%%%%%%%%%%%%%%%%%%%%%%%%%%%%%

\begin{table*}[h]

\begin{subtable}[h]{\textwidth}
\footnotesize
\centering

% Vertical and Horizontal Padding
\setlength{\tabcolsep}{0.5em}
\def\arraystretch{1.5}

\begin{tabular}{L{2cm} C{2cm} C{2cm} C{2cm} C{2cm} C{2.5cm}}
\toprule

\multirow{2}{2cm}{Augmentation Method} & \multicolumn{4}{c}{Learned Impact on Blood Glucose} & \multirow{2}{2.5cm}{\makecell{Learned Dynamics \\ Error}} \\

& 1 hour & 2 hour & 3 hour & 4 hour &\\

\midrule

None & \(0.25 \pm 0.11\) &  \(-0.08 \pm 0.04\) &  \(-0.01 \pm 0.02\) & \(0.08 \pm 0.03\) & \(12.9 \pm 0.6\) \\

Filtering  & \(0.32 \pm 0.12\) &  \(0.33 \pm 0.14\) &  \(-0.27 \pm 0.06\) & \(-0.66 \pm 0.05\) & \(20.2 \pm 2.2\) \\

\textbf{Relabelling} & \textbf{0.10 $\pm$ 0.05} & \textbf{0.41 $\pm$ 0.11} &  \textbf{0.16 $\pm$ 0.07} & \textbf{0.25 $\pm$ 0.04 } & \textbf{9.8 $\pm$ 1.6}\\

\bottomrule
\end{tabular}
\label{tab:carb_comp}
\caption{Carbohydrate's impact on blood glucose following meal consumption.}
\end{subtable}

\vspace{0.5cm}

\begin{subtable}[h]{\textwidth}
\footnotesize
\centering

% Vertical and Horizontal Padding
\setlength{\tabcolsep}{0.5em}
\def\arraystretch{1.5}

\begin{tabular}{L{2cm} C{2cm} C{2cm} C{2cm} C{2cm} C{2.5cm}}
\toprule

\multirow{2}{2cm}{Augmentation Method} & \multicolumn{4}{c}{Learned Impact on Blood Glucose} & \multirow{2}{2.5cm}{\makecell{Learned Dynamics \\ Error}} \\

& 1 hour $\dagger$ & 2 hour & 3 hour & 4 hour &\\

\midrule

None & \(0.11 \pm 0.12\) & \(-0.03 \pm 0.02\) & \(0.02 \pm 0.01\) & \(0.16 \pm 0.02\) & \(32.4 \pm 0.5\) \\

\textbf{Filtering} & \textbf{-0.11 $\pm$ 0.12} &  \textbf{-0.22 $\pm$ 0.01} &  \textbf{-0.28 $\pm$ 0.01} & \textbf{-0.19 $\pm$ 0.02} & \textbf{23.6 $\pm$ 0.7}\\

Relabelling & \(-0.10 \pm 0.13\) & \(-0.10 \pm 0.03\) &  \(-0.10 \pm 0.08\) & \(0.41 \pm 0.04\) & \(29.3 \pm 1.2\) \\

\bottomrule
\end{tabular}
\label{tab:insulin_comp}
\caption{Insulin's impact on blood glucose following subcutaneous insulin dosing.}
    
\end{subtable}

\caption{The learned blood glucose dynamics of the deep learning model trained on augmented data. Filtering removes samples occurring on days with erroneously low carbohydrate consumption and relabelling adds unreported carbohydrate quantities using the mathematical Hovorka model. The learned impact describes the mean insulin or carbohydrates effect on blood glucose after insulin dosing or carbohydrate consumption. The learned dynamics error summarises the total difference between the learned and theoretical blood glucose impact. Augmentation improves the overall correctness of the learned dynamics and reverses the carbohydrate impact such that it is now correctly positive and the insulin impact is correctly negative. Bold highlights the augmentation with the lowest dynamic time warping error. $\dagger$ indicates the results in which the ranking is not statistically significant \((p > 0.05)\) as determined via Friedman rank test and uncertainty is given by the standard error.}
\label{tab:augment_comp}
\end{table*}

%%%%%%%%%%%%%%%%%%%%%%%%%%%%%%%%%%%%%%%%%%%%%%%%%%%%%%%%%%%%%%

% General discussion about filtering and relabelling improving dynamics -> relabelling is slightly more successful
To correct for the conflation of insulin-carbohydrate relationships, relabelling and filtering were applied to the processed dataset. This focused specifically on confusion caused by incorrect or missing meal carbohydrates and was performed to identify if simple modifications could be used to improve blood glucose dynamics without imposing rigid constraints on the deep learning model. Table \ref{tab:augment_comp} summarises the learned dynamics using the described dataset modifications. By comparing the learned dynamics error for carbohydrate consumption, it is clear relabelling improved the learned carbohydrate dynamics of the deep learning simulator. This is evidenced by the reduction in the dynamics error of \(3.3 \pm 2.2\). The simulator also correctly understands that carbohydrate consumption causes future blood glucose levels to rise, as evidenced by the reversal of the learned blood glucose impact values from negative to positive. Similar results can be observed in the learned insulin dynamics with filtering and relabelling both improving the dynamics error by \(8.8 \pm 1.2\) and \(3.1 \pm 1.7\) respectively, in addition to reversing insulin's impact on blood glucose to purely negative across the full four hour period.

% Performance of relabelling
Of the two presented methods for improving learned blood glucose dynamics, the relabelling approach appears to be the most successful, yielding a reduction in the dynamics error for both insulin and carbohydrates. This augmentation method can also be seen to minimise the sharp peak in carbohydrate impact observed 10 minutes post-meal and invert the trough in carbohydrate impact at 95 minutes. The method does deteriorate in the four hour interval, which may suggest that more data is necessary to distinguish the subtle effects of insulin over very long horizons. Despite the improved performance, the learned model is still considerably different from the theoretical glucose dynamics, suggesting there are still confounding factors present in the dataset. 

% Failings of the filtering method
In contrast, the filtering method does improve the learned insulin dynamics, but causes the carbohydrate dynamics to worsen overall, particularly due to reversal in the impact of carbohydrates three to four hours post-meal. This could feasibly also result from difficulty in modelling long-horizon relationships, but may also be a failing of the filtering approach. The method relies on the principle that samples occurring on days where daily carbohydrate consumption falls outside the participant-specific interquartile range are more likely to be affected by unannounced meals. However, a low daily carbohydrate consumption may not always result from an unannounced meal and may simply suggest that participants have opted for a series of low carbohydrate meals on a particular day or that they might be unwell. Incorrectly removing samples would feasibly reduce performance as larger dataset size is strongly linked to improved deep learning predictions.   

% Start peak remains -> another argument for it being related to correct boluses
Despite the observed dynamics improvement with the presented dataset augmentations, the five minute peak in insulin impact remains for both approaches. This observation would support the hypothesis that this error arises  from the high correlation between insulin dosing and carbohydrate consumption, as the filtering and relabelling methods would not alter the timing of these events. In order to correct for this occurrence, it may be necessary to incorporate insulin and CGM sensor delays into the features of the deep learning model. This would clearly indicate the duration after which insulin is likely to affect blood glucose dynamic and could lessen the tight coupling between insulin and carbohydrates. Insulin delay could be incorporated in the simplest case by adding the insulin contribution at the point of greatest insulin impact, as opposed to reporting at the time of infusion as is currently done in the dataset. Similarly, the measurement delay duration should be obtainable from the CGM device's manual and could be included by shifting blood glucose measurements forward by the delay duration. 

\subsubsection{iii) Self-Reported Labels are Ineffective for Modelling High Fat Meals and Physical Activity}

% General results of adding meal and exercise tags
Including labels of exercise and high fat meal consumption as inputs did not notably improve the prediction performance of the deep learning model. The addition of self-reported labels resulted in a slight worsening in the median error from \(1.9 \pm 0.5\) mg/dl to \(2.4 \pm 0.5\) mg/dl. Similarly, the error experienced specifically during high fat meals and exercise events also increased from \(2.4 \pm 0.4\) mg/dl to \(3.4 \pm 0.7\) mg/dl and \(1.6 \pm 0.7\) mg/dl to \(2.8 \pm 0.7\) mg/dl. Wilcoxon signed rank tests across 100 test scenarios did not confirm the difference to be significant (\(p > 0.05\)). The most probable explanation is that the self-reported tags do not provide enough information to make any significant modifications to glucose dynamics. Prior work utilising meal composition data to improve deep glucose forecasting achieved success by combining precise nutritional food values with a compartmental model of glucose absorption \cite{Karim2020After-mealTraining}. This may imply information beyond binary labels of high fat meal consumption are necessary for modelling these events with deep learning simulators. Similarly, glucose levels during physical activity have been observed to be dependent on exercise intensity \cite{Shetty2016EffectDiabetes}, therefore differentiating exercise events in this manner may also be important for effectively describing physical activity. 

% Addition of intensity-grading
Sufficient information was not available in the qualitative labels to add greater detail regarding high fat meal consumption, as the tags mostly neglected to describe the quantity of food being consumed. However, exercise intensity could be feasibly assessed based on descriptions of the physical activity being performed. To incorporate this new information, the binary channel describing exercise in the feature space was split into two sub-channels indicating high and low intensity. High intensity exercise was described as any exercise event deemed to be more strenuous than walking. This division was chosen based on the detail of the information available in the self-reported tags and the assumption that walking would be considered light exercise for most participants.  This distinction resulted in \(20\%\) of exercise events being classified as high intensity and the remaining low intensity. Evaluation of the re-trained deep learning simulator demonstrated that distinguishing exercise events based on intensity also did not significantly improve the modelling of exercise-related events, with the median prediction error remaining unchanged from \(2.4 \pm 0.7\) mg/dl. When distinguishing the results by exercise intensity, low intensity exercise events achieved a reduced error of \(2.7 \pm 0.2\) mg/dl compared to \(2.9 \pm 0.2\) mg/dl for high intensity events. This difference may reflect the larger sample of low intensity events available in the training dataset. 

% Justifying lack of improvement
The lack of improvement observed when grading exercise by intensity could suggest other factors may be limiting deep learning simulation performance. A simple explanation of the plateau in performance could be there are not enough demonstrations of exercise events labelled within the processed dataset. This experiment was performed using data extracted from a single participant, therefore expanding to a larger sample size could feasibly improve modelling predictions. However, the single participant utilised did not record an inconsequential amount of data; with over 1.5 years of blood glucose measurements and a mean of four labelled daily exercise events. This may suggest a very large or detailed dataset would be required for accurately modelling exercise events. Another potential explanation for the minimal improvements in model performance could be that there are already indicators of exercise and high fat meal events visible to the deep learning model. T1D management guidelines often recommend delivering dual-wave boluses to control for high fat meal consumption \cite{Metwally2021InsulinReview}. This control strategy is distinct from a typical meal dose, as a lesser quantity of insulin is given immediately for the consumed carbohydrates and is compensated for by delivering the remaining quantity over a period of several hours. Similarly, exercise is typically accounted for by reducing insulin infusion prior to event occurrence \cite{Thabit2016BasalSpot}. Both these changes could feasibly be visible to the deep learning model provided it had gained an understanding of the typical basal and bolus insulin requirements for a participant.  

\section{Discussion}

% Make actionable recommendations on using free-living data
This work presents a series of actionable recommendations for building deep forecasting algorithms that are more robust to the inherent features likely to be present in real-world diabetes data and for extracting richer and more easily utilisable participant data. These recommendations are broadly applicable to any healthcare domain in which deep learning algorithms are used to model biological systems.

\begin{itemize}

    \item \textbf{Standard Machine Learning metrics are not sufficient for evaluating deep learning simulators.} The deep learning approach was shown to outperform the hybrid and mathematical forecasting algorithms in terms of prediction error, despite conflating the fundamental effects of insulin and carbohydrates on blood glucose dynamics. The vast majority of previously presented deep forecasting algorithms used metrics similar to gMSE as their sole performance measure and therefore could also feasibly be influenced by insulin-carbohydrate confusion \cite{Zhu2021DeepReview,Woldaregay2019Data-drivenDiabetesc}. Novel performance metrics drawing comparisons to known T1D dynamics relationships may be effective in addressing this oversight. Similarly, methods aimed at improving model explainability would further help to identify counter-intuitive or potentially harmful learned relationships.    
    
    \item \textbf{Advanced methods are needed to incorporate expert knowledge without limiting model complexity.} The difficulty of deep learning in modelling insulin-carbohydrate dynamics suggests that it may be necessary to build expert knowledge into simulators to ensure strong performance in safety critical scenarios. The relabelling and filtering dataset augmentations presented in this work did improve the correctness of the learned dynamics, however these approaches only focused on unlabelled meal consumption and therefore performance may be improved further by targeting other features of data collection in diabetes, for example, incorporating the theoretical delay in carbohydrate and insulin impact. Curriculum learning has been applied across other healthcare domains \cite{Wang2022AModels,Ahmed2023GraphClassification} and may be a promising approach for addressing this challenge, as it would allow algorithms to be trained on simpler, more fundamental relationships first before progressing to more subtle effects. 

    \item \textbf{Uncertainty modelling is needed to accurately evaluate high-risk scenarios.} The deep learning model achieved the greatest variance for meals, high-fat meals, and low blood glucose events and struggled to leverage self-reported patient labels to improve meal and exercise modeling. Validating performance in high-risk scenarios, such as these, is essential for ensuring the safety of deep simulation methods, as incorrect predictions may lead to the approval dangerous blood glucose controllers. Poor performance in these events may result from higher measurement uncertainty, caused by factors such as carbohydrate estimation errors or unreported meals misleading the deep learning model. Improved data collection methods could be employed to minimize error, such as by measuring meal composition via image recognition technology \cite{Alfonsi2020CarbohydrateTrial} or passively recording accelerometer data from smartwatches or mobile phones. However due to the inherent nature of uncertainty in real-world data sources, integrating probabilistic deep learning techniques may be a better approach \cite{Zhu2022PersonalizedMeta-Learning,Langarica2023AUncertainties}, as this would enable uncertainty to captured and quantified by the model.

\end{itemize}

\section{Conclusion}

This work explores the challenges of applying deep learning to blood glucose dynamics modelling in T1D. The presented analysis highlights its potential for simulating biological systems; demonstrating notable error reductions over a widely-used mathematical simulator across a range of common blood glucose control scenarios. However, deep learning methods were also observed to be susceptible to misleading features of real-world data, such as highly correlated variables, unreported values, and variables with substantial statistical uncertainty. In some instances this resulted in the confusion of fundamental principles of T1D management, such as the effects of insulin and carbohydrates. Furthermore, these factors impeded the algorithms ability to make accurate predictions in safety critical scenarios and in correctly modelling complex blood glucose events in the presence of self-reported labels, such as high fat meals and exercise. Addressing the above concerns is an important step in integrating these algorithms safely within the current workflow of blood glucose controller development and evaluation. Without guarantees on the physiological appropriateness of the learned blood glucose dynamics, algorithms could leverage dangerously incorrect relationships when making predictions and consequently recommend diabetes management strategies which could harm real-world patients.

This work highlights the need for deep learning evaluation metrics beyond those typically used in machine learning, particularly when training models on real-world data samples to perform safety-critical tasks. These metrics should consider the physiological appropriateness of model predictions and compare them to relationships identified through real-world experimentation. This work presented a simple method of using SHAP values to model insulin and carbohydrate effects, however more sophisticated methods will be required to consider intra-variable interactions and multi-variate prediction tasks. The development of robust evaluation metrics will be important for achieving regulatory approval in deep learning based simulators in T1D and healthcare more broadly, as they provide an interpretable overview of the algorithms expected behaviour in a number of relevant control scenarios and can be used to build trust in model predictions. 

\section{Author Contributions}

H.E., R.M. and M.G. conceptualised the methods and analysis performed in the manuscript. H.E. contributed to the data analysis and acquisition. All authors contibuted to interpretation of the results and edited the manuscript.

\section{Acknowledgements}

This work was supported by the EPSRC Digital Health and Care Centre for Doctoral Training (CDT) at the University of Bristol (UKRI grant no. EP/S023704/1). The authors thank Dana Lewis for providing access to the OpenAPS Data Commons and Enrico Werner and Sam James for their useful discussions.

\section{Competing Interests}

All authors declare no financial or non-financial competing interests. 

\section{Data Availability}

The datasets analysed during the current study are available in the OpenAPS Data Commons repository on reasonable request, (\href{https://openaps.org/outcomes/data-commons/}{https://openaps.org/outcomes/data-commons/}).

\section{Code Availability}

The underlying code for this study is available via GitHub and can be accessed via this link \href{https://github.com/hemerson1/OpenAPS_Cleaner}{https://github.com/hemerson1/OpenAPS\_Cleaner}. 

%% The file named.bst is a bibliography style file for BibTeX 0.99c
\bibliographystyle{naturemag}
\bibliography{references.bib}

\begin{thebibliography}{10}
\expandafter\ifx\csname url\endcsname\relax
  \def\url#1{\texttt{#1}}\fi
\expandafter\ifx\csname urlprefix\endcsname\relax\def\urlprefix{URL }\fi
\providecommand{\bibinfo}[2]{#2}
\providecommand{\eprint}[2][]{\url{#2}}

\bibitem{Leelarathna2021Hybrid2021c}
\bibinfo{author}{Leelarathna, L.} \emph{et~al.}
\newblock \bibinfo{title}{{Hybrid closed-loop therapy: Where are we in 2021?}}
\newblock \emph{\bibinfo{journal}{Diabetes, Obesity and Metabolism}}
  \textbf{\bibinfo{volume}{23}}, \bibinfo{pages}{655--660}
  (\bibinfo{year}{2021}).

\bibitem{Kesavadev2020TheReview}
\bibinfo{author}{Kesavadev, J.}, \bibinfo{author}{Srinivasan, S.},
  \bibinfo{author}{Saboo, B.}, \bibinfo{author}{Krishna, M.} \&
  \bibinfo{author}{Krishnan, G.}
\newblock \bibinfo{title}{{The Do-It-Yourself Artificial Pancreas: A
  Comprehensive Review}}.
\newblock \emph{\bibinfo{journal}{Diabetes Therapy}}
  \textbf{\bibinfo{volume}{11}}, \bibinfo{pages}{1217--1235}
  (\bibinfo{year}{2020}).

\bibitem{Moon2021CurrentEvidence}
\bibinfo{author}{Moon, S.~J.}, \bibinfo{author}{Jung, I.} \&
  \bibinfo{author}{Park, C.~Y.}
\newblock \bibinfo{title}{{Current advances of artificial pancreas systems: A
  comprehensive review of the clinical evidence}}.
\newblock \emph{\bibinfo{journal}{Diabetes and Metabolism Journal}}
  \textbf{\bibinfo{volume}{45}}, \bibinfo{pages}{813--839}
  (\bibinfo{year}{2021}).

\bibitem{Tejedor2020ReinforcementReviewb}
\bibinfo{author}{Tejedor, M.}, \bibinfo{author}{Woldaregay, A.~Z.} \&
  \bibinfo{author}{Godtliebsen, F.}
\newblock \bibinfo{title}{{Reinforcement learning application in diabetes blood
  glucose control: A systematic review}}.
\newblock \emph{\bibinfo{journal}{Artificial Intelligence in Medicine}}
  \textbf{\bibinfo{volume}{104}} (\bibinfo{year}{2020}).

\bibitem{Smaoui2020DevelopmentAlgorithms}
\bibinfo{author}{Smaoui, M.~R.}, \bibinfo{author}{Rabasa-Lhoret, R.} \&
  \bibinfo{author}{Haidar, A.}
\newblock \bibinfo{title}{{Development platform for artificial pancreas
  algorithms}}.
\newblock \emph{\bibinfo{journal}{PLoS ONE}} \textbf{\bibinfo{volume}{15}}
  (\bibinfo{year}{2020}).

\bibitem{Blauw2016}
\bibinfo{author}{Blauw, H.}, \bibinfo{author}{van Bon, A.~C.},
  \bibinfo{author}{Koops, R.} \& \bibinfo{author}{DeVries, J.~H.}
\newblock \bibinfo{title}{{Performance and safety of an integrated bihormonal
  artificial pancreas for fully automated glucose control at home}}
  (\bibinfo{year}{2016}).

\bibitem{Fritzen2018ModelingImpact}
\bibinfo{author}{Fritzen, K.}, \bibinfo{author}{Heinemann, L.} \&
  \bibinfo{author}{Schnell, O.}
\newblock \bibinfo{title}{{Modeling of Diabetes and Its Clinical Impact}}.
\newblock \emph{\bibinfo{journal}{Journal of Diabetes Science and Technology}}
  \textbf{\bibinfo{volume}{12}}, \bibinfo{pages}{976--984}
  (\bibinfo{year}{2018}).

\bibitem{Nath2018PhysiologicalReview}
\bibinfo{author}{Nath, A.}, \bibinfo{author}{Biradar, S.},
  \bibinfo{author}{Balan, A.}, \bibinfo{author}{Dey, R.} \&
  \bibinfo{author}{Padhi, R.}
\newblock \bibinfo{title}{{Physiological Models and Control for Type 1 Diabetes
  Mellitus: A Brief Review}}.
\newblock In \emph{\bibinfo{booktitle}{5th IFAC Conference on Advances in
  Control and Optimization of Dynamical Systems ACODS}},
  vol.~\bibinfo{volume}{51}, \bibinfo{pages}{289--294}
  (\bibinfo{publisher}{Elsevier B.V.}, \bibinfo{year}{2018}).

\bibitem{Bhonsle2020AStrategies}
\bibinfo{author}{Bhonsle, S.} \& \bibinfo{author}{Saxena, S.}
\newblock \bibinfo{title}{{A review on control-relevant glucose–insulin
  dynamics models and regulation strategies}}.
\newblock In \emph{\bibinfo{booktitle}{Proceedings of the Institution of
  Mechanical Engineers. Part I: Journal of Systems and Control Engineering}},
  vol. \bibinfo{volume}{234}, \bibinfo{pages}{596--608}
  (\bibinfo{publisher}{SAGE Publications Ltd}, \bibinfo{year}{2020}).

\bibitem{Woldaregay2019Data-drivenDiabetesc}
\bibinfo{author}{Woldaregay, A.~Z.} \emph{et~al.}
\newblock \bibinfo{title}{{Data-driven modeling and prediction of blood glucose
  dynamics: Machine learning applications in type 1 diabetes}}
  (\bibinfo{year}{2019}).

\bibitem{Porumb2020PrecisionECG}
\bibinfo{author}{Porumb, M.}, \bibinfo{author}{Stranges, S.},
  \bibinfo{author}{Pescap{\`{e}}, A.} \& \bibinfo{author}{Pecchia, L.}
\newblock \bibinfo{title}{{Precision Medicine and Artificial Intelligence: A
  Pilot Study on Deep Learning for Hypoglycemic Events Detection based on
  ECG}}.
\newblock \emph{\bibinfo{journal}{Scientific Reports}}
  \textbf{\bibinfo{volume}{10}} (\bibinfo{year}{2020}).

\bibitem{Cescon2021ActivityConditions}
\bibinfo{author}{Cescon, M.} \emph{et~al.}
\newblock \bibinfo{title}{{Activity detection and classification from wristband
  accelerometer data collected on people with type 1 diabetes in free-living
  conditions}}.
\newblock \emph{\bibinfo{journal}{Computers in Biology and Medicine}}
  \textbf{\bibinfo{volume}{135}} (\bibinfo{year}{2021}).

\bibitem{Zhao2020TrainingApplications}
\bibinfo{author}{Zhao, Q.}, \bibinfo{author}{Adeli, E.} \&
  \bibinfo{author}{Pohl, K.~M.}
\newblock \bibinfo{title}{{Training confounder-free deep learning models for
  medical applications}}.
\newblock \emph{\bibinfo{journal}{Nature Communications}}
  \textbf{\bibinfo{volume}{11}} (\bibinfo{year}{2020}).

\bibitem{Zhu2022PersonalizedMeta-Learning}
\bibinfo{author}{Zhu, T.}, \bibinfo{author}{Li, K.}, \bibinfo{author}{Herrero,
  P.} \& \bibinfo{author}{Georgiou, P.}
\newblock \bibinfo{title}{{Personalized Blood Glucose Prediction for Type 1
  Diabetes Using Evidential Deep Learning and Meta-Learning}}.
\newblock \emph{\bibinfo{journal}{IEEE Transactions on Biomedical Engineering}}
   (\bibinfo{year}{2022}).

\bibitem{Kushner2020Multi-HourModels-correct}
\bibinfo{author}{Kushner, T.}, \bibinfo{author}{Breton, M.~D.} \&
  \bibinfo{author}{Sankaranarayanan, S.}
\newblock \bibinfo{title}{{Multi-Hour Blood Glucose Prediction in Type 1
  Diabetes: A Patient-Specific Approach Using Shallow Neural Network Models}}.
\newblock \emph{\bibinfo{journal}{Diabetes Technology and Therapeutics}}
  \textbf{\bibinfo{volume}{22}}, \bibinfo{pages}{883--891}
  (\bibinfo{year}{2020}).

\bibitem{Hovorka2004NonlinearDiabetes}
\bibinfo{author}{Hovorka, R.} \emph{et~al.}
\newblock \bibinfo{title}{{Nonlinear model predictive control of glucose
  concentration in subjects with type 1 diabetes}}.
\newblock In \emph{\bibinfo{booktitle}{Physiological Measurement}},
  vol.~\bibinfo{volume}{25}, \bibinfo{pages}{905--920}
  (\bibinfo{publisher}{Institute of Physics Publishing}, \bibinfo{year}{2004}).

\bibitem{Pompa2021ASystem}
\bibinfo{author}{Pompa, M.}, \bibinfo{author}{Panunzi, S.},
  \bibinfo{author}{Borri, A.} \& \bibinfo{author}{de~Gaetano, A.}
\newblock \bibinfo{title}{{A comparison among three maximal mathematical models
  of the glucose-insulin system}}.
\newblock \emph{\bibinfo{journal}{PLoS ONE}} \textbf{\bibinfo{volume}{16}}
  (\bibinfo{year}{2021}).

\bibitem{Panunzi2020ALoad}
\bibinfo{author}{Panunzi, S.}, \bibinfo{author}{Pompa, M.},
  \bibinfo{author}{Borri, A.}, \bibinfo{author}{Piemonte, V.} \&
  \bibinfo{author}{de~Gaetano, A.}
\newblock \bibinfo{title}{{A revised Sorensen model: Simulating glycemic and
  insulinemic response to oral and intra-venous glucose load}}.
\newblock \emph{\bibinfo{journal}{PLoS ONE}} \textbf{\bibinfo{volume}{15}}
  (\bibinfo{year}{2020}).

\bibitem{Visentin2018TheDayb}
\bibinfo{author}{Visentin, R.} \emph{et~al.}
\newblock \bibinfo{title}{{The UVA/Padova Type 1 Diabetes Simulator Goes From
  Single Meal to Single Day}}.
\newblock \emph{\bibinfo{journal}{Journal of Diabetes Science and Technology}}
  \textbf{\bibinfo{volume}{12}}, \bibinfo{pages}{273--281}
  (\bibinfo{year}{2018}).

\bibitem{Alkhateeb2021ModellingDiabetes}
\bibinfo{author}{Alkhateeb, H.}, \bibinfo{author}{El~Fathi, A.},
  \bibinfo{author}{Ghanbari, M.} \& \bibinfo{author}{Haidar, A.}
\newblock \bibinfo{title}{{Modelling glucose dynamics during moderate exercise
  in individuals with type 1 diabetes}}.
\newblock \emph{\bibinfo{journal}{PLoS ONE}} \textbf{\bibinfo{volume}{16}}
  (\bibinfo{year}{2021}).

\bibitem{Resalat2019AModel}
\bibinfo{author}{Resalat, N.}, \bibinfo{author}{Youssef, J.~E.},
  \bibinfo{author}{Tyler, N.}, \bibinfo{author}{Castle, J.} \&
  \bibinfo{author}{Jacobs, P.~G.}
\newblock \bibinfo{title}{{A statistical virtual patient population for the
  glucoregulatory system in type 1 diabetes with integrated exercise model}}.
\newblock \emph{\bibinfo{journal}{PLoS ONE}} \textbf{\bibinfo{volume}{14}}
  (\bibinfo{year}{2019}).

\bibitem{Romeres2021ExerciseStudy}
\bibinfo{author}{Romeres, D.} \emph{et~al.}
\newblock \bibinfo{title}{{Exercise effect on insulin-dependent and
  insulin-independent glucose utilization in healthy individuals and
  individuals with type 1 diabetes: A modeling study}}.
\newblock \emph{\bibinfo{journal}{American Journal of Physiology -
  Endocrinology and Metabolism}} \textbf{\bibinfo{volume}{321}},
  \bibinfo{pages}{E122--E129} (\bibinfo{year}{2021}).

\bibitem{Deng2022Patient-specificDiabetes}
\bibinfo{author}{Deng, Y.}, \bibinfo{author}{Arao, K.},
  \bibinfo{author}{Mantzoros, C.~S.} \& \bibinfo{author}{Karniadakis, G.~E.}
\newblock \bibinfo{title}{{Patient-specific deep offline artificial pancreas
  for blood glucose regulation in type 1 diabetes}}.
\newblock \emph{\bibinfo{journal}{BioArxiv}}  (\bibinfo{year}{2022}).
\newblock \urlprefix\url{https://doi.org/10.1101/2022.10.21.513303}.

\bibitem{Yazdani2020SystemsDynamics}
\bibinfo{author}{Yazdani, A.}, \bibinfo{author}{Lu, L.},
  \bibinfo{author}{Raissi, M.} \& \bibinfo{author}{Karniadakis, G.~E.}
\newblock \bibinfo{title}{{Systems biology informed deep learning for inferring
  parameters and hidden dynamics}}.
\newblock \emph{\bibinfo{journal}{PLoS Computational Biology}}
  \textbf{\bibinfo{volume}{16}} (\bibinfo{year}{2020}).

\bibitem{Zhu2021DeepReview}
\bibinfo{author}{Zhu, T.}, \bibinfo{author}{Li, K.}, \bibinfo{author}{Herrero,
  P.} \& \bibinfo{author}{Georgiou, P.}
\newblock \bibinfo{title}{{Deep Learning for Diabetes: A Systematic Review}}.
\newblock \emph{\bibinfo{journal}{IEEE Journal of Biomedical and Health
  Informatics}} \textbf{\bibinfo{volume}{25}}, \bibinfo{pages}{2744--2757}
  (\bibinfo{year}{2021}).

\bibitem{Ahmed2023TheReview}
\bibinfo{author}{Ahmed, A.} \emph{et~al.}
\newblock \bibinfo{title}{{The Effectiveness of Wearable Devices Using
  Artificial Intelligence for Blood Glucose Level Forecasting or Prediction:
  Systematic Review}}.
\newblock \emph{\bibinfo{journal}{Journal of Medical Internet Research}}
  \textbf{\bibinfo{volume}{25}}, \bibinfo{pages}{e40259}
  (\bibinfo{year}{2023}).

\bibitem{Li2020GluNet:Forecasting}
\bibinfo{author}{Li, K.}, \bibinfo{author}{Liu, C.}, \bibinfo{author}{Zhu, T.},
  \bibinfo{author}{Herrero, P.} \& \bibinfo{author}{Georgiou, P.}
\newblock \bibinfo{title}{{GluNet: A Deep Learning Framework for Accurate
  Glucose Forecasting}}.
\newblock \emph{\bibinfo{journal}{IEEE Journal of Biomedical and Health
  Informatics}} \textbf{\bibinfo{volume}{24}}, \bibinfo{pages}{414--423}
  (\bibinfo{year}{2020}).

\bibitem{Fox2018DeepForecastingb}
\bibinfo{author}{Fox, I.}, \bibinfo{author}{Ang, L.}, \bibinfo{author}{Jaiswal,
  M.}, \bibinfo{author}{Pop-Busui, R.} \& \bibinfo{author}{Wiens, J.}
\newblock \bibinfo{title}{{Deep Multi-Output Forecasting}}.
\newblock In \emph{\bibinfo{booktitle}{Proceedings of the 24th ACM SIGKDD
  International Conference on Knowledge Discovery {\&} Data Mining}}
  (\bibinfo{year}{2018}).

\bibitem{Zhu2020BloodNetworks}
\bibinfo{author}{Zhu, T.}, \bibinfo{author}{Yao, X.}, \bibinfo{author}{Li, K.},
  \bibinfo{author}{Herrero, P.} \& \bibinfo{author}{Georgiou, P.}
\newblock \bibinfo{title}{{Blood Glucose Prediction for Type 1 Diabetes Using
  Generative Adversarial Networks}}.
\newblock In \emph{\bibinfo{booktitle}{Proceedings of the 5th International
  Workshop on Knowledge Discovery in Healthcare Data co-located with 24th
  ECAI}} (\bibinfo{year}{2020}).

\bibitem{Miller2020LearningWildb}
\bibinfo{author}{Miller, A.~C.}, \bibinfo{author}{Foti, N.~J.} \&
  \bibinfo{author}{Fox, E.}
\newblock \bibinfo{title}{{Learning Insulin-Glucose Dynamics in the Wild}}.
\newblock \emph{\bibinfo{journal}{Machine Learning for Healthcare}}
  \textbf{\bibinfo{volume}{126}}, \bibinfo{pages}{1--25}
  (\bibinfo{year}{2020}).
\newblock \urlprefix\url{http://arxiv.org/abs/2008.02852}.

\bibitem{Hameed2020ComparingData}
\bibinfo{author}{Hameed, H.} \& \bibinfo{author}{Kleinberg, S.}
\newblock \bibinfo{title}{{Comparing Machine Learning Techniques for Blood
  Glucose Forecasting Using Free-living and Patient Generated Data}}.
\newblock In \emph{\bibinfo{booktitle}{Proceedings of Machine Learning
  Research}}, \bibinfo{pages}{871--894} (\bibinfo{year}{2020}).

\bibitem{Kushner2020ConformanceDynamics}
\bibinfo{author}{Kushner, T.}, \bibinfo{author}{Sankaranarayanan, S.} \&
  \bibinfo{author}{Breton, M.}
\newblock \bibinfo{title}{{Conformance verification for neural network models
  of glucose-insulin dynamics}}.
\newblock In \emph{\bibinfo{booktitle}{HSCC 2020 - Proceedings of the 23rd
  International Conference on Hybrid Systems: Computation and Control ,part of
  CPS-IoT Week}} (\bibinfo{publisher}{Association for Computing Machinery,
  Inc}, \bibinfo{year}{2020}).

\bibitem{Prudencio2023AProblems}
\bibinfo{author}{Prudencio, R.~F.}, \bibinfo{author}{Maximo, M. R. O.~A.} \&
  \bibinfo{author}{Colombini, E.~L.}
\newblock \bibinfo{title}{{A Survey on Offline Reinforcement Learning:
  Taxonomy, Review, and Open Problems}}.
\newblock \emph{\bibinfo{journal}{IEEE Transations on Neural Networks and
  Learning Systems}}  (\bibinfo{year}{2023}).
\newblock \urlprefix\url{http://arxiv.org/abs/2203.01387}.

\bibitem{Fu2021BenchmarksEvaluationb}
\bibinfo{author}{Fu, J.} \emph{et~al.}
\newblock \bibinfo{title}{{Benchmarks for Deep Off-Policy Evaluation}}.
\newblock In \emph{\bibinfo{booktitle}{International Conference on Learning
  Representations (ICLR)}} (\bibinfo{year}{2021}).
\newblock \urlprefix\url{https://github.com/google-research/deep_}.

\bibitem{Cai2023TowardsLearning}
\bibinfo{author}{Cai, X.}, \bibinfo{author}{Chen, J.}, \bibinfo{author}{Zhu,
  Y.}, \bibinfo{author}{Wang, B.} \& \bibinfo{author}{Yao, Y.}
\newblock \bibinfo{title}{{Towards Safe Propofol Dosing during General
  Anesthesia Using Deep Offline Reinforcement Learning}}
  (\bibinfo{year}{2023}).
\newblock \urlprefix\url{http://arxiv.org/abs/2303.10180}.

\bibitem{Shiranthika2022SupervisedLearning}
\bibinfo{author}{Shiranthika, C.} \emph{et~al.}
\newblock \bibinfo{title}{{Supervised Optimal Chemotherapy Regimen Based on
  Offline Reinforcement Learning}}.
\newblock \emph{\bibinfo{journal}{IEEE Journal of Biomedical and Health
  Informatics}} \textbf{\bibinfo{volume}{26}}, \bibinfo{pages}{4763--4772}
  (\bibinfo{year}{2022}).

\bibitem{Kondrup2022TowardsLearning}
\bibinfo{author}{Kondrup, F.} \emph{et~al.}
\newblock \bibinfo{title}{{Towards Safe Mechanical Ventilation Treatment Using
  Deep Offline Reinforcement Learning}}  (\bibinfo{year}{2022}).
\newblock \urlprefix\url{http://arxiv.org/abs/2210.02552}.

\bibitem{Jennings2020Do-It-YourselfProfessionals}
\bibinfo{author}{Jennings, P.} \& \bibinfo{author}{Hussain, S.}
\newblock \bibinfo{title}{{Do-It-Yourself Artificial Pancreas Systems: A Review
  of the Emerging Evidence and Insights for Healthcare Professionals}}.
\newblock \emph{\bibinfo{journal}{Journal of Diabetes Science and Technology}}
  \textbf{\bibinfo{volume}{14}}, \bibinfo{pages}{868--877}
  (\bibinfo{year}{2020}).

\bibitem{Street2021ReviewOutcomes}
\bibinfo{author}{Street, T.~J.}
\newblock \bibinfo{title}{{Review of Self-Reported Data from UK Do-It-Yourself
  Artificial Pancreas System (DIYAPS) Users to Determine Whether Demographic of
  Population Affects Use or Outcomes}}.
\newblock \emph{\bibinfo{journal}{Diabetes Therapy}}
  \textbf{\bibinfo{volume}{12}}, \bibinfo{pages}{1839--1848}
  (\bibinfo{year}{2021}).

\bibitem{Mcknight2015GlycaemicComparison}
\bibinfo{author}{Mcknight, J.~A.} \emph{et~al.}
\newblock \bibinfo{title}{{Glycaemic control of Type 1 diabetes in clinical
  practice early in the 21st century: An international comparison}}.
\newblock \emph{\bibinfo{journal}{Diabetic Medicine}}
  \textbf{\bibinfo{volume}{32}}, \bibinfo{pages}{1036--1050}
  (\bibinfo{year}{2015}).

\bibitem{Lin2019HandlingRecords}
\bibinfo{author}{Lin, K.-P.}, \bibinfo{author}{Magjarevic, R.} \&
  \bibinfo{author}{De~Carvalho, P.}
\newblock \bibinfo{title}{{Handling Missing Data in CGM Records}}.
\newblock In \emph{\bibinfo{booktitle}{International Conference on Biomedical
  and Health Informatics}}, \bibinfo{pages}{420--427} (\bibinfo{year}{2019}).
\newblock \urlprefix\url{http://www.springer.com/series/7403}.

\bibitem{Charlton2020ABeverages}
\bibinfo{author}{Charlton, J.} \emph{et~al.}
\newblock \bibinfo{title}{{A review of the challenges, glycaemic risks and
  self-care for people with type 1 diabetes when consuming alcoholic
  beverages}}.
\newblock \emph{\bibinfo{journal}{Practical Diabetes}}
  \textbf{\bibinfo{volume}{37}}, \bibinfo{pages}{7--12} (\bibinfo{year}{2020}).

\bibitem{Paterson2015TheManagementb}
\bibinfo{author}{Paterson, M.} \emph{et~al.}
\newblock \bibinfo{title}{{The Role of Dietary Protein and Fat in Glycaemic
  Control in Type 1 Diabetes: Implications for Intensive Diabetes Management}}.
\newblock \emph{\bibinfo{journal}{Current Diabetes Reports}}
  \textbf{\bibinfo{volume}{15}} (\bibinfo{year}{2015}).

\bibitem{Paterson2016InfluenceTherapy}
\bibinfo{author}{Paterson, M.~A.} \emph{et~al.}
\newblock \bibinfo{title}{{Influence of dietary protein on postprandial blood
  glucose levels in individuals with Type 1 diabetes mellitus using intensive
  insulin therapy}}.
\newblock \emph{\bibinfo{journal}{Diabetic Medicine}}
  \textbf{\bibinfo{volume}{33}}, \bibinfo{pages}{592--598}
  (\bibinfo{year}{2016}).

\bibitem{Abdou2021EffectEgyptb}
\bibinfo{author}{Abdou, M.} \emph{et~al.}
\newblock \bibinfo{title}{{Effect of high protein and fat diet on postprandial
  blood glucose levels in children and adolescents with type 1 diabetes in
  Cairo, Egypt}}.
\newblock \emph{\bibinfo{journal}{Diabetes and Metabolic Syndrome: Clinical
  Research and Reviews}} \textbf{\bibinfo{volume}{15}}, \bibinfo{pages}{7--12}
  (\bibinfo{year}{2021}).

\bibitem{Dewar2017TheDiabetes}
\bibinfo{author}{Dewar, L.} \& \bibinfo{author}{Heuberger, R.}
\newblock \bibinfo{title}{{The effect of acute caffeine intake on insulin
  sensitivity and glycemic control in people with diabetes}}.
\newblock \emph{\bibinfo{journal}{Diabetes and Metabolic Syndrome: Clinical
  Research and Reviews}} \textbf{\bibinfo{volume}{11}},
  \bibinfo{pages}{S631--S635} (\bibinfo{year}{2017}).

\bibitem{Montville2013USDA5.0}
\bibinfo{author}{Montville, J.~B.} \emph{et~al.}
\newblock \bibinfo{title}{{USDA Food and Nutrient Database for Dietary Studies
  (FNDDS), 5.0}}.
\newblock \emph{\bibinfo{journal}{Procedia Food Science}}
  \textbf{\bibinfo{volume}{2}}, \bibinfo{pages}{99--112}
  (\bibinfo{year}{2013}).

\bibitem{Smart2020InsulinTime}
\bibinfo{author}{Smart, C.~E.}, \bibinfo{author}{King, B.~R.} \&
  \bibinfo{author}{Lopez, P.~E.}
\newblock \bibinfo{title}{{Insulin dosing for fat and protein: Is it time?}}
  (\bibinfo{year}{2020}).

\bibitem{Chang2017DilatedNetworks}
\bibinfo{author}{Chang, S.} \emph{et~al.}
\newblock \bibinfo{title}{{Dilated Recurrent Neural Networks}}.
\newblock In \emph{\bibinfo{booktitle}{Neural Information Processing Systems}}
  (\bibinfo{year}{2017}).
\newblock \urlprefix\url{https://github.com/code-terminator/DilatedRNN}.

\bibitem{Kuroda2012Carbohydrate-to-insulinPump}
\bibinfo{author}{Kuroda, A.} \emph{et~al.}
\newblock \bibinfo{title}{{Carbohydrate-to-insulin ratio is estimated from
  300-400 divided by total daily insulin dose in type 1 diabetes patients who
  use the insulin pump}}.
\newblock \emph{\bibinfo{journal}{Diabetes Technology and Therapeutics}}
  \textbf{\bibinfo{volume}{14}}, \bibinfo{pages}{1077--1080}
  (\bibinfo{year}{2012}).

\bibitem{Favero2012AModels}
\bibinfo{author}{Favero, S.~D.}, \bibinfo{author}{Facchinetti, A.} \&
  \bibinfo{author}{Cobelli, C.}
\newblock \bibinfo{title}{{A glucose-specific metric to assess predictors and
  identify models}}.
\newblock \emph{\bibinfo{journal}{IEEE Transactions on Biomedical Engineering}}
  \textbf{\bibinfo{volume}{59}}, \bibinfo{pages}{1281--1290}
  (\bibinfo{year}{2012}).

\bibitem{Cengiz2016MovingPumpsb}
\bibinfo{author}{Cengiz, E.}, \bibinfo{author}{Bode, B.},
  \bibinfo{author}{Van~Name, M.} \& \bibinfo{author}{Tamborlane, W.~V.}
\newblock \bibinfo{title}{{Moving toward the ideal insulin for insulin pumps}}.
\newblock \emph{\bibinfo{journal}{Expert Review of Medical Devices}}
  \textbf{\bibinfo{volume}{13}}, \bibinfo{pages}{57--69}
  (\bibinfo{year}{2016}).

\bibitem{Sundararajan2017AxiomaticNetworks}
\bibinfo{author}{Sundararajan, M.}, \bibinfo{author}{Taly, A.} \&
  \bibinfo{author}{Yan, Q.}
\newblock \bibinfo{title}{{Axiomatic Attribution for Deep Networks}}.
\newblock In \emph{\bibinfo{booktitle}{International Conference on Machine
  Learning}} (\bibinfo{year}{2017}).

\bibitem{Zheng2019AutomatedExplanationb}
\bibinfo{author}{Zheng, M.}, \bibinfo{author}{Ni, B.} \&
  \bibinfo{author}{Kleinberg, S.}
\newblock \bibinfo{title}{{Automated meal detection from continuous glucose
  monitor data through simulation and explanation}}.
\newblock \emph{\bibinfo{journal}{Journal of the American Medical Informatics
  Association}} \textbf{\bibinfo{volume}{26}}, \bibinfo{pages}{1592--1599}
  (\bibinfo{year}{2019}).

\bibitem{Yu2022DeepDiabetes}
\bibinfo{author}{Yu, X.} \emph{et~al.}
\newblock \bibinfo{title}{{Deep transfer learning: a novel glucose prediction
  framework for new subjects with type 2 diabetes}}.
\newblock \emph{\bibinfo{journal}{Complex and Intelligent Systems}}
  \textbf{\bibinfo{volume}{8}}, \bibinfo{pages}{1875--1887}
  (\bibinfo{year}{2022}).

\bibitem{Briscoe2006HypoglycemiaManagement}
\bibinfo{author}{Briscoe, V.~J.} \& \bibinfo{author}{Davis, S.~N.}
\newblock \bibinfo{title}{{Hypoglycemia in Type 1 and Type 2 Diabetes:
  Physiology, Pathophysiology, and Management}}.
\newblock \emph{\bibinfo{journal}{Clinical Diabetes}}
  \textbf{\bibinfo{volume}{24}}, \bibinfo{pages}{115--121}
  (\bibinfo{year}{2006}).

\bibitem{Meade2016AccuracyAdults}
\bibinfo{author}{Meade, L.~T.} \& \bibinfo{author}{Rushton, W.~E.}
\newblock \bibinfo{title}{{Accuracy of carbohydrate counting in adults}}.
\newblock \emph{\bibinfo{journal}{Clinical Diabetes}}
  \textbf{\bibinfo{volume}{34}}, \bibinfo{pages}{142--147}
  (\bibinfo{year}{2016}).

\bibitem{Smith2021InsulinReview}
\bibinfo{author}{Smith, T.~A.}, \bibinfo{author}{Marlow, A.~A.},
  \bibinfo{author}{King, B.~R.} \& \bibinfo{author}{Smart, C.~E.}
\newblock \bibinfo{title}{{Insulin strategies for dietary fat and protein in
  type 1 diabetes: A systematic review}} (\bibinfo{year}{2021}).

\bibitem{Bequette2013AlgorithmsControl}
\bibinfo{author}{Bequette, B.~W.}
\newblock \bibinfo{title}{{Algorithms for a Closed-Loop Artificial Pancreas:
  The Case for Model Predictive Control}}.
\newblock \bibinfo{type}{Tech. Rep.} \bibinfo{number}{6}
  (\bibinfo{year}{2013}).

\bibitem{Slattery2018OptimalReview}
\bibinfo{author}{Slattery, D.}, \bibinfo{author}{Amiel, S.~A.} \&
  \bibinfo{author}{Choudhary, P.}
\newblock \bibinfo{title}{{Optimal prandial timing of bolus insulin in diabetes
  management: a review}}.
\newblock \emph{\bibinfo{journal}{Diabetic Medicine}}
  \textbf{\bibinfo{volume}{35}}, \bibinfo{pages}{306--316}
  (\bibinfo{year}{2018}).

\bibitem{Akturk2018PossibleDiabetes}
\bibinfo{author}{Akturk, H.~K.}, \bibinfo{author}{Rewers, A.},
  \bibinfo{author}{Joseph, H.}, \bibinfo{author}{Schneider, N.} \&
  \bibinfo{author}{Garg, S.~K.}
\newblock \bibinfo{title}{{Possible Ways to Improve Postprandial Glucose
  Control in Type 1 Diabetes}}.
\newblock \emph{\bibinfo{journal}{Diabetes Technology and Therapeutics}}
  \textbf{\bibinfo{volume}{20}}, \bibinfo{pages}{S224--S232}
  (\bibinfo{year}{2018}).

\bibitem{DenckerJohansen2012InterindividualMellitus}
\bibinfo{author}{Dencker~Johansen, M.}, \bibinfo{author}{Gjerl{\o}v, I.},
  \bibinfo{author}{Sandahl~Christiansen, J.}, \bibinfo{author}{Hejlesen, O.~K.}
  \& \bibinfo{author}{Author, C.}
\newblock \bibinfo{title}{{Interindividual and Intraindividual Variations in
  Postprandial Glycemia Peak Time Complicate Precise Recommendations for
  Self-Monitoring of Glucose in Persons with Type 1 Diabetes Mellitus}}.
\newblock \bibinfo{type}{Tech. Rep.} \bibinfo{number}{2}
  (\bibinfo{year}{2012}).
\newblock \urlprefix\url{www.journalofdst.org}.

\bibitem{Berget2019ATherapy}
\bibinfo{author}{Berget, C.}, \bibinfo{author}{Messer, L.~H.} \&
  \bibinfo{author}{Forlenza, G.~P.}
\newblock \bibinfo{title}{{A clinical overview of insulin pump therapy for the
  management of diabetes: Past, present, and future of intensive therapy}}
  (\bibinfo{year}{2019}).

\bibitem{Wong2021Ultra-rapid-actingNeeded}
\bibinfo{author}{Wong, E.~Y.} \& \bibinfo{author}{Kroon, L.}
\newblock \bibinfo{title}{{Ultra-rapid-acting insulins: How fast is really
  needed?}} (\bibinfo{year}{2021}).

\bibitem{Smartphone-basedScienceDirect}
\bibinfo{title}{{Smartphone-based vs paper-based asthma action plans for
  adolescents - ScienceDirect}}.

\bibitem{Nohara2022ExplanationHospital}
\bibinfo{author}{Nohara, Y.}, \bibinfo{author}{Matsumoto, K.},
  \bibinfo{author}{Soejima, H.} \& \bibinfo{author}{Nakashima, N.}
\newblock \bibinfo{title}{{Explanation of machine learning models using shapley
  additive explanation and application for real data in hospital}}.
\newblock \emph{\bibinfo{journal}{Computer Methods and Programs in
  Biomedicine}} \textbf{\bibinfo{volume}{214}} (\bibinfo{year}{2022}).

\bibitem{Cappon2020ADiabetes}
\bibinfo{author}{Cappon, G.} \emph{et~al.}
\newblock \bibinfo{title}{{A Personalized and Interpretable Deep Learning Based
  Approach to Predict Blood Glucose Concentration in Type 1 Diabetes}}.
\newblock In \emph{\bibinfo{booktitle}{Knowledge Discovery in Healthcare Data
  at the European Conference on Artificial Intelligence}}
  (\bibinfo{year}{2020}).

\bibitem{Mase2019ExplainingRefinement}
\bibinfo{author}{Mase, M.}, \bibinfo{author}{Owen, A.~B.} \&
  \bibinfo{author}{Seiler, B.}
\newblock \bibinfo{title}{{Explaining black box decisions by Shapley cohort
  refinement}} (\bibinfo{year}{2019}).
\newblock \urlprefix\url{http://arxiv.org/abs/1911.00467}.

\bibitem{Karim2020After-mealTraining}
\bibinfo{author}{Karim, R.~A.}, \bibinfo{author}{Vass{\'{a}}nyi, I.} \&
  \bibinfo{author}{K{\'{o}}sa, I.}
\newblock \bibinfo{title}{{After-meal blood glucose level prediction using an
  absorption model for neural network training}}.
\newblock \emph{\bibinfo{journal}{Computers in Biology and Medicine}}
  \textbf{\bibinfo{volume}{125}} (\bibinfo{year}{2020}).

\bibitem{Shetty2016EffectDiabetes}
\bibinfo{author}{Shetty, V.~B.} \emph{et~al.}
\newblock \bibinfo{title}{{Effect of exercise intensity on glucose requirements
  to maintain euglycemia during exercise in type 1 diabetes}}.
\newblock \emph{\bibinfo{journal}{Journal of Clinical Endocrinology and
  Metabolism}} \textbf{\bibinfo{volume}{101}}, \bibinfo{pages}{972--980}
  (\bibinfo{year}{2016}).

\bibitem{Metwally2021InsulinReview}
\bibinfo{author}{Metwally, M.}, \bibinfo{author}{Cheung, T.~O.},
  \bibinfo{author}{Smith, R.} \& \bibinfo{author}{Bell, K.~J.}
\newblock \bibinfo{title}{{Insulin pump dosing strategies for meals varying in
  fat, protein or glycaemic index or grazing-style meals in type 1 diabetes: A
  systematic review}}.
\newblock \emph{\bibinfo{journal}{Diabetes Research and Clinical Practice}}
  \textbf{\bibinfo{volume}{172}} (\bibinfo{year}{2021}).

\bibitem{Thabit2016BasalSpot}
\bibinfo{author}{Thabit, H.} \& \bibinfo{author}{Leelarathna, L.}
\newblock \bibinfo{title}{{Basal insulin delivery reduction for exercise in
  type 1 diabetes: finding the sweet spot}}.
\newblock \emph{\bibinfo{journal}{Diabetologia}} \textbf{\bibinfo{volume}{59}},
  \bibinfo{pages}{1628--1631} (\bibinfo{year}{2016}).

\bibitem{Wang2022AModels}
\bibinfo{author}{Wang, Y.}, \bibinfo{author}{Han, X.}, \bibinfo{author}{Hao,
  X.}, \bibinfo{author}{Zhu, T.} \& \bibinfo{author}{Shu, H.}
\newblock \bibinfo{title}{{A Curriculum Batching Strategy for Automatic ICD
  Coding with Deep Multi-Label Classification Models}}.
\newblock \emph{\bibinfo{journal}{Healthcare (Basel)}}
  \textbf{\bibinfo{volume}{10}} (\bibinfo{year}{2022}).

\bibitem{Ahmed2023GraphClassification}
\bibinfo{author}{Ahmed, U.}, \bibinfo{author}{Lin, J. C.~W.} \&
  \bibinfo{author}{Srivastava, G.}
\newblock \bibinfo{title}{{Graph Attention-based Curriculum Learning for Mental
  Healthcare Classification}}.
\newblock \emph{\bibinfo{journal}{IEEE Journal of Biomedical and Health
  Informatics}}  (\bibinfo{year}{2023}).

\bibitem{Alfonsi2020CarbohydrateTrial}
\bibinfo{author}{Alfonsi, J.~E.} \emph{et~al.}
\newblock \bibinfo{title}{{Carbohydrate counting app using image recognition
  for youth with Type 1 diabetes: Pilot randomized control trial}}.
\newblock \emph{\bibinfo{journal}{JMIR mHealth and uHealth}}
  \textbf{\bibinfo{volume}{8}} (\bibinfo{year}{2020}).

\bibitem{Langarica2023AUncertainties}
\bibinfo{author}{Langarica, S.}, \bibinfo{author}{Rodriguez-Fernandez, M.},
  \bibinfo{author}{Doyle, F.~J.} \& \bibinfo{author}{Nunez, F.}
\newblock \bibinfo{title}{{A Probabilistic Approach to Blood Glucose Prediction
  in Type 1 Diabetes Under Meal Uncertainties}}.
\newblock \emph{\bibinfo{journal}{IEEE Journal of Biomedical and Health
  Informatics}}  (\bibinfo{year}{2023}).

\end{thebibliography}

\end{document}